\documentclass[10pt,twocolumn,letterpaper]{article}

\usepackage[final,pagenumbers]{cvpr}
\usepackage[utf8]{inputenc} 
\usepackage{graphicx}
\usepackage{amsmath}
\usepackage{amssymb}
\usepackage{booktabs}
\usepackage{times}
\usepackage{epsfig} 
\usepackage{makecell}
\usepackage{times}
\usepackage{silence}
\usepackage{epstopdf}
\usepackage{color, colortbl}
\usepackage{float}  
\usepackage{caption}
\usepackage{multirow} 
\usepackage{multicol}
\usepackage{xcolor}
\usepackage{url}
\usepackage{placeins}
\usepackage[group-separator={,}]{siunitx}
\usepackage[utf8]{inputenc}
\usepackage{amssymb}
\usepackage{pifont}
\usepackage{paralist}
\usepackage{floatpag}

\usepackage{tcolorbox}
\usepackage{listings}
\usepackage{xcolor}
\usepackage{inconsolata} 

\definecolor{jsonkeypink}{RGB}{219, 48, 122}
\definecolor{jsoncomment}{RGB}{0, 110, 0}    
\definecolor{jsonnumber}{RGB}{30, 144, 255}

\newcommand{\xmark}{\ding{55}}

\newcommand{\vspaceundertab}{\vspace{-.2cm}}

\definecolor{LightCyan}{rgb}{0.88,1,1}
\definecolor{Gray}{gray}{0.9}
\definecolor{Pink}{rgb}{1, 0, 1}
\definecolor{azure}{rgb}{0.0, 0.44, 1.0}
\definecolor{bleudefrance}{rgb}{0.19, 0.55, 0.91}
\definecolor{cobalt}{rgb}{0.0, 0.28, 0.67}
\definecolor{electricpurple}{rgb}{0.75, 0.0, 1.0}

\newcommand{\na}{\gr{N/A}}
\newcommand{\sft}{\az{\textbf{SFT}}\xspace}
\newcommand{\rl}{\teal{\textbf{RL}}\xspace}

\newcommand{\sftt}{\az{\textbf{[+SFT]}}\xspace}
\newcommand{\rlt}{\teal{\textbf{[+RL]}}\xspace}

\definecolor{ai2green}{HTML}{105357}

\newcommand{\direct}{{\colorbox{gray!20!white}{\textsf{\textsc{Direct}}}}}
\newcommand{\agent}{{\colorbox{teal!20!white}{\textsf{\textsc{Agent}}}}}

\newcommand{\re}[1]{\textcolor{red}{#1}}

\newcommand{\gr}[1]{\textcolor{gray}{#1}}

\newcommand{\az}[1]{\textcolor{azure}{#1}}

\newcommand{\teal}[1]{\textcolor{teal}{#1}}
\definecolor{LightCyan}{rgb}{0.88,1,1}

\newcommand{\system}{SAGE\xspace}
\newcommand{\model}{SAGE-MM\xspace}
\newcommand{\benchmark}{SAGE-Bench\xspace}
\definecolor{cvprblue}{rgb}{0.21,0.49,0.74}
\usepackage[pagebackref,breaklinks,colorlinks,allcolors=cvprblue,urlcolor=magenta]{hyperref}

\usepackage[capitalize]{cleveref}
\crefname{section}{Sec.}{Secs.}
\Crefname{section}{Section}{Sections}
\Crefname{table}{Table}{Tables}
\crefname{table}{Tab.}{Tabs.}

\def\confName{CVPR}
\def\confYear{2026}

\begin{document}

\title{\system: Training Smart Any-Horizon Agents for Long Video Reasoning with Reinforcement Learning}

\author{
  Jitesh Jain\textsuperscript{1,2}\thanks{Work done during JJ's internship at Allen AI. \quad \textsuperscript{$\dagger$}Equal advising.} \quad
  Jialuo Li\textsuperscript{1} \quad
  Zixian Ma\textsuperscript{2,3} \quad
  Jieyu Zhang\textsuperscript{2,3} \quad 
  Chris Dongjoo Kim\textsuperscript{2} \quad Sangho Lee\textsuperscript{2} \quad \\[0.2em]
  Rohun Tripathi\textsuperscript{2} \quad 
  Tanmay Gupta\textsuperscript{2} \quad Christopher Clark\textsuperscript{2}\textsuperscript{$\dagger$} \quad
  Humphrey Shi\textsuperscript{1}\textsuperscript{$\dagger$} \\[0.4em]
  \textsuperscript{1}SHI Labs @ Georgia Tech \quad 
  \textsuperscript{2}Allen AI \quad 
  \textsuperscript{3}University of Washington \\[0.4em]
  \textbf{\texttt{\url{https://github.com/allenai/SAGE}}}
}

\maketitle

\begin{abstract}

\noindent
As humans, we are natural any-horizon reasoners, \textit{i.e.}, we can decide whether to iteratively skim long videos or watch short ones in full when necessary for a given task. With this in mind, one would expect video reasoning models to reason flexibly across different durations. However, SOTA models are still trained to predict answers in a single turn while processing a large number of frames, akin to watching an entire long video, requiring significant resources. This raises the question: \textbf{Is it possible to develop performant any-horizon video reasoning systems?}
Inspired by human behavior, we first propose \textbf{\system}, an agent system that performs multi-turn reasoning on long videos while handling simpler problems in a single turn. Secondly, we introduce an easy synthetic data generation pipeline using Gemini-2.5-Flash to train the orchestrator, \textbf{\model}, which lies at the core of \system. We further propose an effective RL post-training recipe essential for instilling any-horizon reasoning ability in \model. Thirdly, we curate \textbf{\benchmark} with an average duration of greater than 700 seconds for evaluating video reasoning ability in real-world entertainment use cases. Lastly, we empirically validate the effectiveness of our system, data, and RL recipe, observing notable improvements of up to \textbf{6.1\%} on open-ended video reasoning tasks, as well as an impressive \textbf{8.2\%} improvement on videos longer than 10 minutes. 

\end{abstract}
\vspace{-0.3cm}
\section{Introduction}

\begin{figure}[t!]
\centering
\includegraphics[width=1\linewidth]{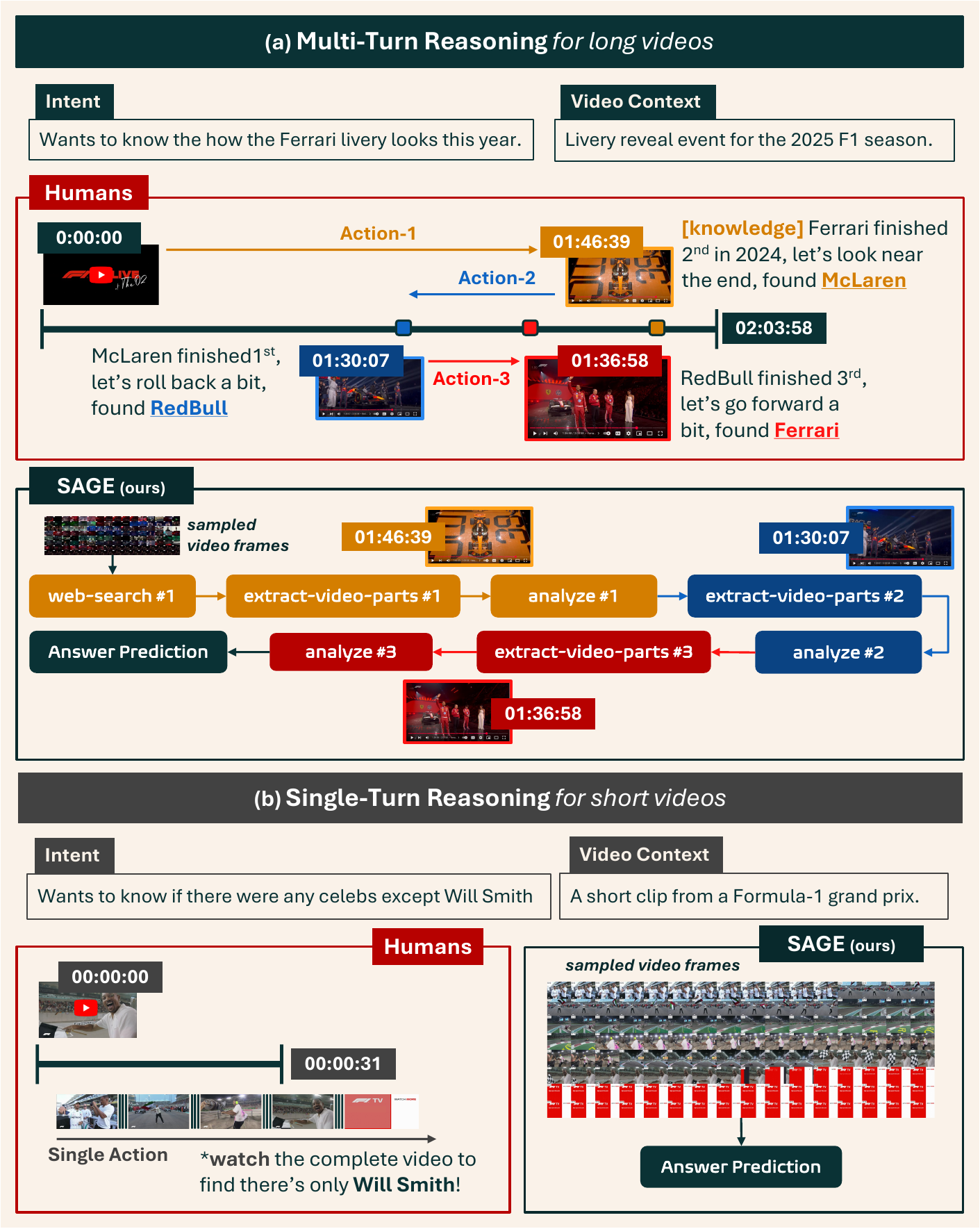} \\
\vspace{-0.2cm}
\caption{\textbf{Human behavior-inspired design of \system.} We design \system to resemble humans' adaptive reasoning behavior, capable of following a knowledge-driven multi-turn reasoning process using tool calls for long-horizon tasks~(\cref{tab:tool_set}) while being able to predict an answer for short-horizon problems directly.}
\vspace{-0.4cm}
\label{fig:teaser}
\end{figure}

\noindent
In the last year, there has been a natural shift from developing models for solely image reasoning~\cite{liu2023improvedllava, tong2024cambrian1, zhu2023minigpt, molmo2024, lin2023vila, jain2025visper_lm, jain2024vcoder, chen2023internvl, chen2024expanding, Qwen2-VL} to also tackling video reasoning~\cite{zhu2025internvl3, chen2024longvila, Qwen2.5-VL, wang2025internvl3_5, qwen3-vl, gemini25pushingfrontier} in the research community. Among the various model releases, the recent Gemini-2.5~\cite{gemini25pushingfrontier} and Qwen3-VL~\cite{qwen3-vl} models pushed the frontier in video reasoning due to their ability to perform well on both short and long videos. 

Although the aforementioned SOTA models differ in their training data, recipe, and architecture, among other things, they all function in a standard way when reasoning over videos: given a set of sampled frames, output the final answer with a single sequence prediction process, \emph{i.e.}, single turn reasoning. We refer to this line of work as falling under the {\direct} paradigm. Orthogonal to the works mentioned above, a few methods~\cite{liu2025videomind, wang2025videochat, zhi2025videoagent2enhancingllmbasedagent, chen2025lvagent, li2024wolf, goldfish} take an agentic route to predicting answers through multi-turn reasoning, falling under the {\agent} paradigm.

Humans excel at tasks that require multi-turn reasoning. For example, when viewing a 2-hour-long video, as humans, we take an iterative approach to finding the target information (\cref{fig:teaser}).
With the recent overwhelming success of RL post-training for training multi-turn agent systems for long-horizon tasks like software engineering~\cite {qwen3technicalreport, yang2024sweagent, faircodegenteam2025cwmopenweightsllmresearch}, computer-use~\cite{yang2025gta1guitesttimescaling, seed2025seed1_5vl, qwen3-vl}, and deep-research~\cite{dong2025tool, deepagent, tongyidr}, it is natural to expect multi-turn agent systems to do well at long video reasoning. Despite the analogy above, most of the existing long video reasoning systems are still trained following the {\direct} paradigm, even with RL~\cite{chen2025longvila-r1, wang2025internvl3_5}.

Motivated by the above realization, we explore the question: \textbf{What are the technical challenges toward effectively training video reasoning models under the {\agent} paradigm with Reinforcement Learning?} We outline three significant aspects for answering the above question: training data \textbf{(A1)}, efficient system design \textbf{(A2)}, and RL recipe for multi-turn reasoning \textbf{(A3)}.

\textbf{(A1)} The training data for an agent model capable of long video reasoning requires access to high-quality question-answer (QnA) pairs. Collecting QnA pairs for long videos poses a daunting challenge due to their lengthy duration. For example, having a human annotate a single 1-hour-long video can cost approximately \$30 on the Prolific platform, making it expensive for data collection at scale. To avoid such high costs, existing works typically employ a synthetic data curation process by iteratively processing 10-30 second-long subclips using models adept at short video understanding to either generate QnA pairs directly~\cite{eagle25} or captions followed by QnA pairs using an LLM~\cite{chen2024longvila, chen2025longvila-r1}. Although inexpensive compared to human annotation, the mentioned bottom-up pipeline is slow and resource-intensive — imagine processing 120 subclips for an hour-long video; even with each subclip taking only 10 seconds, it would take 20 minutes to process a single video. Therefore, to save time and money, we leverage the long-context modeling capabilities of Gemini-2.5-Flash to generate synthetic, high-quality QnA pairs with a carefully designed prompt, ensuring the generated questions span the whole video. Moreover, we manually verify over 1700 generated samples and find a low 5\% error rate while achieving nearly 100$\times$ cost and 10$\times$ time savings compared to human annotation and subclip processing pipelines, respectively.

\textbf{(A2)} Existing multi-turn agent systems usually use an LLM/VLM to orchestrate the calls to only a temporal grounder tool~\cite{liu2025videomind, videoexplorer, dang2025mupamultipathagenticreasoning} to iteratively locate an event over the entire video needed for finding an answer to a given question. However, we posit that attempting to ground an event in the whole video is not always the most effective approach due to the lack of robust temporal grounding models for long videos.
For example, knowing the Formula 1 2024 season standings enables intelligent reasoning with a small temporal search space when watching the 2025 season livery reveal event video (\cref{fig:teaser}\textcolor{cvprblue}{a}).  Motivated by similar use cases, we introduce the \textbf{\system} (\textbf{S}mart \textbf{A}ny-horizon a\textbf{GE}nt) system for long video reasoning. Particularly, we take a more innovative approach by equipping our system with tools such as web search and speech transcription, in addition to temporal grounding, to ensure that it is adept at not only utilizing visual signals from the video but also leveraging verbal and external knowledge. At the core of our system lies an orchestrator VLM, \textbf{\model}, responsible for deciding between multi-turn and single-turn behavior for effective any-horizon reasoning.
Moreover, guided by the fact that a user typically interacts with videos for entertainment~\cite{hafuta2025video, dannenbaum2025five}, we focus our efforts on verifying the effectiveness of our approach on \textbf{\benchmark}, curated with videos from popular YouTube channels to simulate use cases in the daily lives of users. Interestingly, we find existing agent systems to be over-engineered toward answering multiple-choice questions, often underperforming at the open-ended problems under \benchmark (\cref{tab:results}), demonstrating their ineffectiveness for real-world use-cases.

\textbf{(A3)} The variable duration of videos presents a unique challenge to training multi-turn agents. Specifically, during the RL post-training stage, the model should learn to function as an any-horizon agent, i.e., directly output the answer for simple problems while using multi-turn reasoning for harder problems~\cite{zhan2025katv1kwaiautothinktechnicalreport}. We believe that the optimization challenge posed by the dynamic nature of videos presents a challenge for training agent models using existing RL recipes, which have been shown to work well for training {\direct} models~\cite{chen2025longvila-r1, video-r1}. Moreover, extending the RLVR techniques~\cite{deepseekr1, deepseek-math} to video reasoning presents another challenge due to the task's open-ended nature, which results in a lack of verifiable rewards. A few {\direct} approaches~\cite{wang2025video, chen2025longvila-r1} overcome the verifiable reward challenge by training only on MCQ problems and/or using some form of string-overlap metrics~\cite{video-r1, VideoRFT}, rendering them ineffective at open-ended problems (\cref{tab:results}).

\begin{figure*}[ht!]
\centering
\includegraphics[width=1\linewidth]{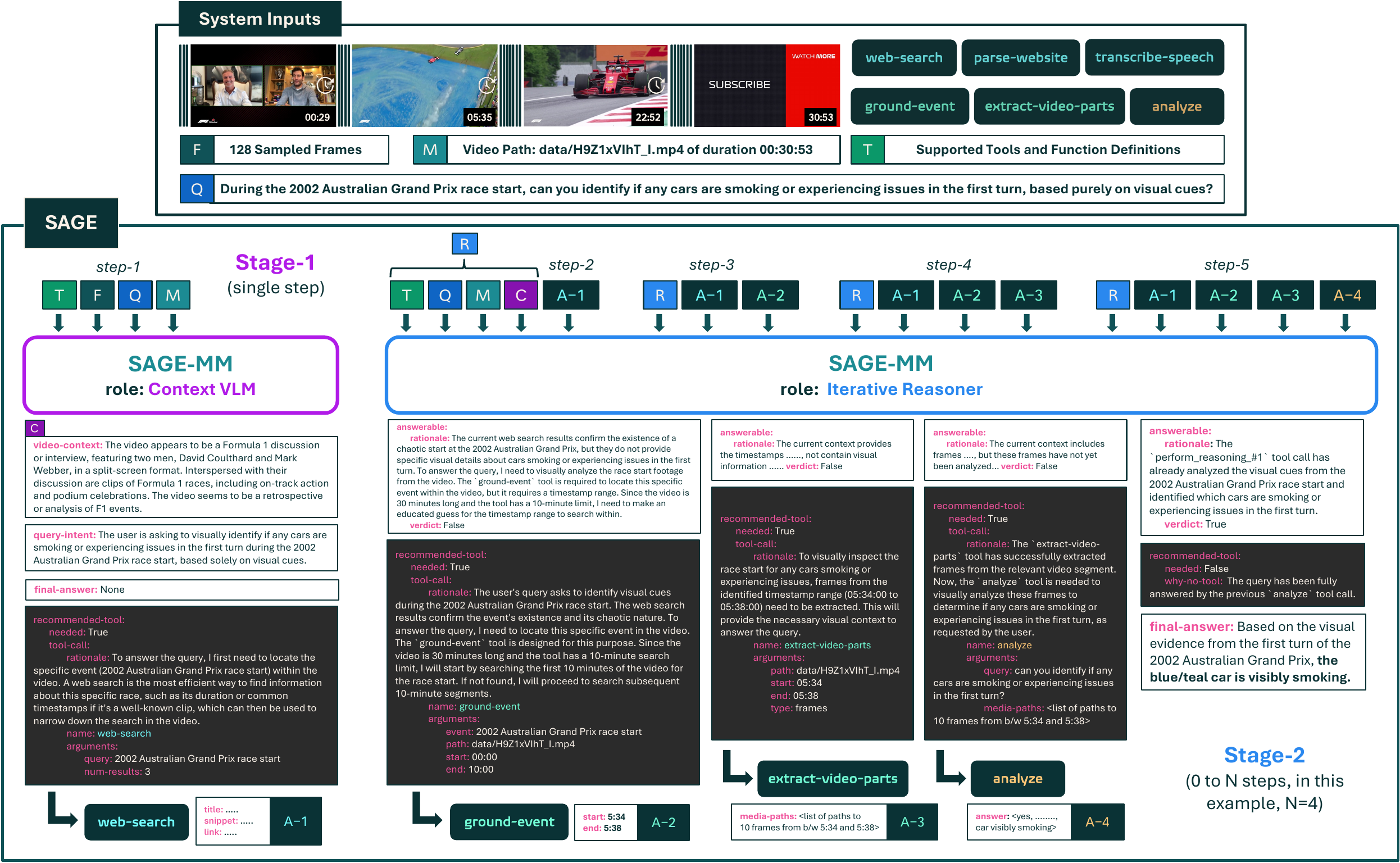} \\
\vspace{-0.2cm}
\caption{\textbf{\system Workflow.} Our system accepts four inputs (shown at the top): sampled video frames ($F$), metadata about the video ($M$), available tool definitions ($T$), and the user query ($Q$). Given these inputs, \system operates in two stages based on the role of \model. In \textbf{Stage-1}, \model is responsible for providing information about the video's context ($C$) along with either a final answer prediction or a tool call to be executed before the next step. At every subsequent step in \textbf{Stage-2}, \model uses the video context ($C$) and the tool call results from previous steps to decide either to predict the final answer or call another tool in an iterative reasoning process.}
\vspace{-0.4cm}
\label{fig:sage}
\end{figure*}

To that end, we propose a multi-reward RL recipe that utilizes strong reasoning LLMs~\cite{openai2024gpt4ocard} to validate the correctness of answers during the RL post-training stage. Moreover, moving away from using string-matching for evaluation, we adopt a universal LLM-as-a-judge evaluation approach to maintain uniformity across our training and evaluation setups. Our RL recipe improves the SFT model by 4.1\% and surpasses the base by 5.7\%, demonstrating its effectiveness. Moreover, for videos longer than 10 minutes, we observe performance improvements of up to \textbf{14.6\%} along with \textbf{4.8\%} for videos shorter than 10 minutes, proving \system's effectiveness on any-horizon video reasoning.

In summary, we make the following contributions:

\begin{compactitem}
    \item We propose \textbf{\system}, an any-horizon agent for long-video reasoning, equipped with a web-search tool for knowledge-driven multi-turn reasoning.
    \item We introduce a cost-effective synthetic QnA pipeline using Gemini-2.5-Flash to train and evaluate our system on entertainment videos for real-world use.
    \item We train \model with an effective RL post-training recipe to instill any-horizon reasoning, demonstrating the scalability of our system design for RL.
\end{compactitem}
\section{Related Work}

\begin{table*}[t!]
  \centering
  \fontsize{10}{12}\selectfont
  \resizebox{1.\linewidth}{!}{
  \begin{tabular}{l|l|l|l}

 tool-name & purpose & arguments & returns  \\
\midrule

\textbf{\texttt{web-search}} & Perform web search using a text query. & query (\textit{str}); num-results (\textit{int}) & List of URL, title, and snippet for search results. \\

\textbf{\texttt{parse-website}} & Parse web data from a given URL. & website-url (\textit{str}) & Parsed HTML content of the website. \\

\textbf{\texttt{transcribe-speech}} & Perform ASR on the video. & path (\textit{str}), start (\textit{str}), end (\textit{str}) & Segment-level verbal transcript between the start and end timestamps. \\

\textbf{\texttt{ground-event}} & Identify timestamps for an event in the video. & event (\textit{str}), path (\textit{str}), start (\textit{str}), end (\textit{str}) & Timestamps for the event between the start and end timestamps. \\

\textbf{\texttt{extract-video-parts}} & Extract frames or subclips between two timestamps. & type (\textit{str}), path (\textit{str}), start (\textit{str}), end (\textit{str}) & List of paths to the saved extracted parts (either frames or a subclip). \\

\textbf{\texttt{analyze}} & Analyze a set of media based on a query. &  query (\textit{str}), media-paths (\textit{List[str]}) & Answer to the query. \\

\bottomrule
\end{tabular}
}
  \vspaceundertab
  \caption{\textbf{Supported tools in \system.} Our system has access to six tools, including web search (via the Serper-hosted Google Search API), for performing knowledge-driven reasoning. We implement the ground-event and analyze tools using existing MLLMs~\cite{qwen3-vl}.}
    \label{tab:tool_set}
    \vspace{-0.3cm}
\end{table*}

\subsection{Long Video Reasoning Agents}

\noindent

Existing long video reasoning agent systems are usually composed of two core components: \textit{an orchestrator}, and \textit{a tool set}, with a temporal grounder being a standard tool among all methods. The orchestrator is responsible for determining the actions to execute while interacting with the available tools within a multi-turn pipeline. A common aspect in the design of existing long video reasoning agent systems is their over-reliance on a temporal grounding module to perform event-guided multi-turn reasoning. 

VideoAgent~\cite{fan2025videoagent} creates a memory using the caption and keyframe features from the video subclips and incorporates tools to retrieve information from memory for reasoning. Similarly, VideoChat-A1~\cite{wang2025videochat} employs keyframe retrieval to perform chain-of-shot reasoning. VideoMind~\cite{liu2025videomind} tunes LoRA adapters for the base Qwen2-VL~\cite{Qwen2-VL} model as a verifier to verify outputs from a separate temporal grounder module before final answer prediction. VideoExplorer~\cite{videoexplorer} optimizes the planner module with DPO~\cite{rafailov2023direct} for better trajectory reasoning. LVAgent~\cite{chen2025lvagent} leverages collaboration among multiple MLLMs with iterative reflection and key frame perception to reach the final answer.

In this work, we move away from over-reliance on temporal grounding and incorporate tools such as web search and speech transcription to enable an intelligent event localization strategy. Moreover, unlike the above methods, \system attempts to predict timestamps for an event within one short subclip at a time rather than the entire video, based on probable coarse event boundaries generated by \model, resulting in a more efficient approach.

\subsection{Reinforcement Learning for Video Reasoning}

\noindent
Following the success of DeepSeek-R1~\cite{deepseekr1} at using Reinforcement Learning with Verifiable Rewards (RLVR) to improve reasoning abilities in LLMs, various works have tried to leverage GRPO~\cite{deepseek-math, deepseekr1} to train {\direct} video reasoning models capable of \textit{thinking} and then answering. Video-R1~\cite{video-r1} follows the optimization approach of DeepSeek-R1 and introduces a contrastive temporal variant of GRPO, comparing answers between inputs with correct and incorrect frame ordering to enforce temporal dependence during reasoning. VideoRFT~\cite{VideoRFT} introduces a semantic-consistency reward between the reasoning trace and video frames. Video-Thinker~\cite{wang2025video} optimizes the model to output multiple temporal grounding instances within a single reasoning trace by carefully curating the cold-start SFT dataset. LongVILA-R1~\cite{chen2025longvila-r1} enables the use of thousands of frames during the RL post-training stage with sequence parallelism. All the above methods utilize option-matching and ROUGE metrics to compute rewards, rendering their approach suboptimal for open-ended problems.

We train \model to learn the ability to perform any-horizon reasoning using GRPO while leveraging an LLM-as-a-Judge to handle rewards for open-ended problems. A concurrent work, LongVT~\cite{yang2025longvt} also employs a similar training recipe with LLM-as-a-judge for computing the accuracy reward while only supporting a crop-video tool call.
\section{Method}

\begin{figure*}[t!]
\centering
\includegraphics[width=1\linewidth]{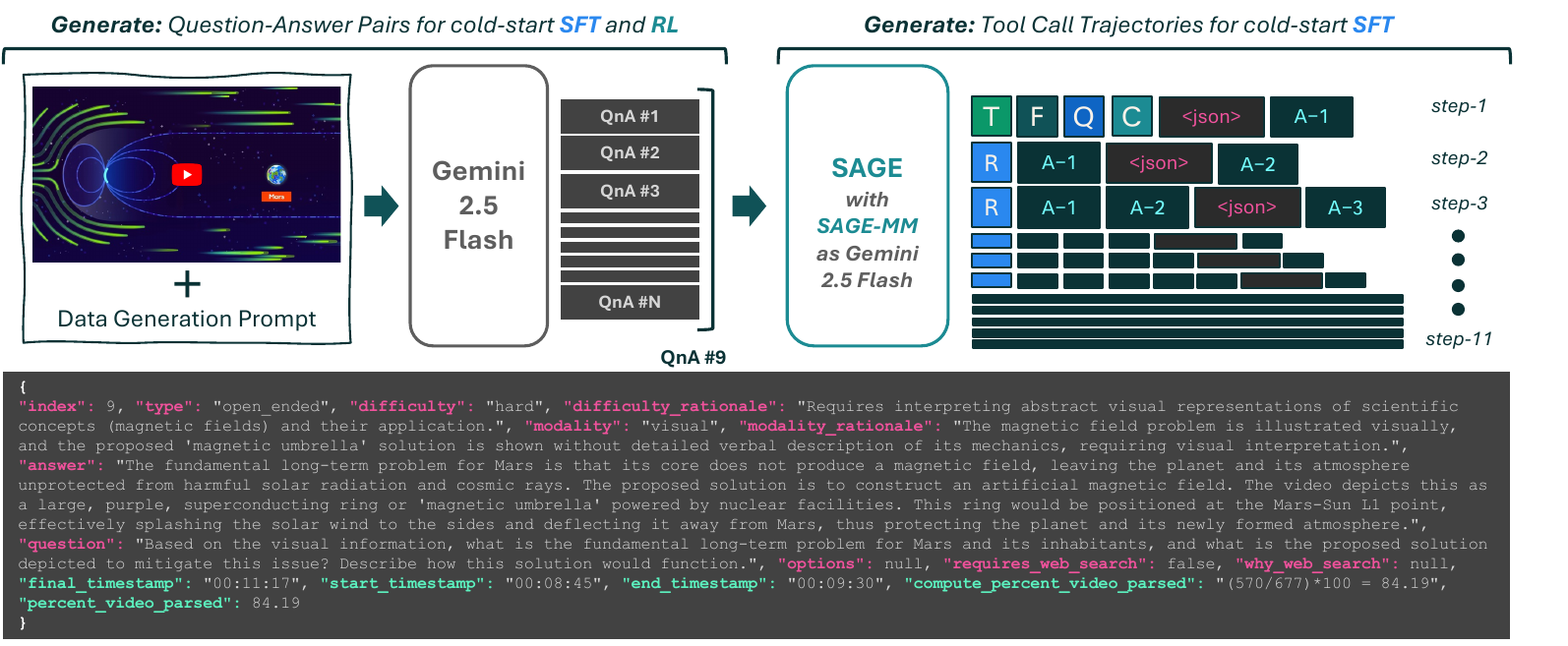} \\
\vspace{-0.3cm}
\caption{\textbf{Synthetic Data Generation Pipeline.} We leverage Gemini-2.5-Flash to generate 10-20 QnA pairs, covering the full temporal span of the video. We find that instructing the model to predict a \textbf{\texttt{percent\_video\_parsed}} field for every QnA pair helps in enforcing proper coverage. We use a \system with Gemini-2.5-Flash as the orchestrator to synthesize tool call trajectories for a cold-start SFT stage.}
\vspace{-0.4cm}
\label{fig:data}
\end{figure*}

\noindent
In the daily life of a human, entertainment is the primary purpose for interacting with videos~\cite{dannenbaum2025five, hafuta2025video}, from watching sports videos on YouTube to scrolling through hundreds of short reels on Instagram. Therefore, it's only natural to develop video reasoning models, keeping the user's needs in mind. Among those needs, the open-ended interaction holds a vital place. For instance, as shown in \cref{fig:teaser}, a user would usually ask: \textit{"How does the Ferrari livery look this year?"} as an open-ended question and expect the model to provide an answer in real-time. We introduce \textbf{\system}, a system designed to answer users' questions while they enjoy entertainment videos. In the following subsections, we present technical details about \system (\cref{subsec:sys}), followed by our synthetic data generation pipeline (\cref{subsec:data}). Lastly, we provide information on training the orchestrator (\model) using RL for the system (\cref{subsec:train}). 

\subsection{System Design}
\label{subsec:sys}

\noindent
As shown at the top of \cref{fig:sage}, our \system expects four inputs: 128 sampled frames from the video ($F$), metadata about the video ($M$), available tools' definitions ($T$), and the user query ($Q$). \system operates in two stages, based on the role of the orchestrator (\model) (\cref{fig:sage} bottom):

\vspace{0.1em}
\noindent
\textbf{Stage-1 (role: Context VLM):} In this single-step stage, \model accepts the system inputs ($T$\textbar$F$\textbar$Q$\textbar$M$) and outputs a JSON action string with required fields:

\setlength{\leftmargini}{0pt}
\begin{compactitem}
    \item \underline{\textit{video-context}} ($C$): Information about the video's setting.
    \item \underline{\textit{query-intent}}: The intent behind the user's query. 
    \item \underline{\textit{recommended-tool}}: Information about the next tool call if a final answer cannot be generated at the current step.
    \item \underline{\textit{final-answer}}: null if tool call; otherwise predicted answer.
\end{compactitem}

\vspace{0.1em}
\noindent
The metadata string ($M$) comprises information about the video path and duration, which are necessary to predict the arguments for the tool call. We list the supported tools in \system in \cref{tab:tool_set}. Notably, unlike previous methods, which either perform temporal grounding over the complete video~\cite{liu2025videomind, videoexplorer}, our \system autonomously predicts segment-level timestamps to ground events over a maximum duration of 10 minutes, as we qualitatively found that existing models struggle on longer entertainment videos.

\vspace{0.1em}
\noindent
\textbf{Stage-2 (role: Iterative Reasoner):} In this multi-step stage, \model accepts the tool call and video context results from all the previous steps, along with the other textual inputs ($T$\textbar$Q$\textbar$M$) and decides if the user query can be answered or another tool call is needed. At every step, \model outputs a JSON action string with three required fields:

\setlength{\leftmargini}{0pt}
\begin{compactitem}
    \item \underline{\textit{answerable}}: Whether the query can be answered. 
    \item \underline{\textit{recommended-tool}}: Information about the next tool call if a final answer cannot be generated at the current step.
    \item \underline{\textit{final-answer}}: null if tool call; otherwise predicted answer.
\end{compactitem}

\vspace{0.1em}
\noindent
We set the maximum number of steps under stage 2 to ten to prevent indefinite execution length. We provide an example execution graph for \system at the bottom of \cref{fig:sage}.

\subsection{Synthetic Data Generation}
\label{subsec:data}

\noindent
We collect videos and shorts from 13 popular YouTube channels across diverse genres, including sports (\textit{Formula1}), food (\textit{ZachChoi}), comedy (\textit{TheDailyShow, MrBean, TheOffice, Friends, fluffyguy, trevornoah}), education (\textit{Vox, kurzgesagt, veritasium, QuantaScienceChannel}), and travel (\textit{WalkingAlice}). Given a video, our synthetic data generation pipeline includes two stages: (i) question-answer (QnA) pair generation using Gemini-2.5-Flash for training and evaluation, and (ii) tool call trajectory generation using \system with Gemini-2.5-Flash as the \model for cold-start SFT, as shown in \cref{fig:data}.

\noindent
\textbf{QnA Pairs.} We leverage the long context modeling abilities of Gemini-2.5-Flash~\cite{gemini25pushingfrontier} to generate questions and answers for a given video in a single pass using a carefully designed prompt. We find that for videos longer than 5 minutes, having the model predict a \textbf{\texttt{percent\_video\_parsed}} field is critical to ensure that the generated questions temporally span the complete video, as shown at the bottom of \cref{fig:data}. We generate 10-20 QnA pairs per video.

\noindent
\textbf{Tool Call Trajectories.} We observe that existing open-source VLMs are not adept at functioning as \model right off the shelf, which is a necessity for successful RL post-training. Therefore, we also generate four tool call trajectories for each question and use input-action pairs from unique trajectories to create a cold-start SFT dataset to finetune our own \model model before the RL post-training stage. \cref{tab:data_stats} lists statistics for our training data.

\subsection{RL Post Training}
\label{subsec:train}

\noindent
We use GRPO~\cite{deepseek-math, deepseekr1} as the policy optimization algorithm during the RL post-training stage for trajectory-level optimization. Specifically, during the rollout generation, the $i^{th}$ action rollout trajectory for a given input set $S_1$ = $\{T, F, M, Q\}$ is represented by $\tau_{i}$. Therefore, we can formulate $\tau_{i}$ as a sequence of state-action pairs $\forall j \in [0, N]$:

\begin{table}[t!]
  \centering
  \fontsize{10}{12}\selectfont
  \resizebox{1.\linewidth}{!}{
  \begin{tabular}{l|ccccccc|c}
{} & {0--60} & {60--180} & {180--300} & {300--600} & {600--1200} & {1200--2400} & {2400+} & {total} \\
\midrule

\textbf{\#videos} & 1493 & 1907 & 552 & 612 & 1067 & 461 & 576 & 6668 \\

\textbf{\#QnA} & 20.2k & 23.0k & 8.0k & 9.4k & 22.2k & 7.1k & 9.4k & 99.1k \\

\textbf{\#actions} & 43.4k & 43.9k & 38.6k & 49.6k & 115.0k & 52.2k & 75.0k & 417.7k \\

\bottomrule
\end{tabular}
}
  \vspaceundertab
  \caption{\textbf{Training Data Statistics.} We generate over 99k questions for more than 6600 videos from popular YouTube channels.}
    \label{tab:data_stats}
    \vspace{-0.2cm}
\end{table}

\vspace{-1.0em}
\begin{equation}
\begin{aligned}
& \tau_i = \bigl[(S_1, A_1), (S_2, A_2), \ldots, (S_N, A_{N})\bigr], \\
& A_j = \textbf{SAGE-MM}(S_j), \\
& S_{j+1} = \{T, Q, M, C, A_{1}...A_{j}\}
\end{aligned}
\end{equation}

\noindent
During the advantage computation step in GRPO, we assign a single scalar reward $R_i$
to every action in the trajectory $\tau_i$ with $N$ steps. The reward consists of (i) step-level
rewards $s_j$ collected at each step, and (ii) a final accuracy reward $a_N$ at the
end of the trajectory. The resulting reward $R_i$ is then uniformly assigned to all
actions in $\tau_i$:

\vspace{-0.7em}
\begin{equation}
\begin{aligned}
& R_i = (s_1 + s_2 + s_3 +...+ s_N) \;+\; a_N \\
& r(A_1) = r(A_2) =....= r(A_N) = R_i
\end{aligned}
\end{equation}
\vspace{-0.7em}

\noindent
Note that we can assign final rewards to all steps because rollout generation is synchronous, \emph{i.e.}, advantages are computed only after all trajectories are completed in a batch.

\vspace{0.1em}
\noindent
\textbf{Step-Level Rewards.} The reward ($s_j$) for a step $j$ in a trajectory is a sum of four scores:

\setlength{\leftmargini}{0pt}
\begin{compactitem}
    \item \textit{format}: Encourages producing a JSON action string with only the required fields.
    \vspace{-0.2cm}
    \[
    s_{\text{format}} =
    \begin{cases}
        +0.05, & \text{if JSON contains only required fields} \\
        -0.10, & \text{otherwise}
    \end{cases}
    \]
    \vspace{-0.3cm}

    \item \textit{reasonable-tool}: Encourages the model to perform sensible multi-step tool usage. Specifically, at each step, we ask GPT-4o to judge whether the current tool call is rational, given the question and the previous tool calls.
    
    \vspace{-0.4cm}
    \[
    s_{\text{reasonable-tool}} =
    \begin{cases}
        +0.10, & \text{if current tool call is reasonable} \\
        -0.10, & \text{otherwise}
    \end{cases}
    \]

    \vspace{-0.2em}

    \item \textit{args-repeat}: Penalizes repetitive tool call arguments. 
    \vspace{-0.2em}
    \[
    s_{\text{args-repeat}} = -0.05 \cdot \sqrt{\text{num-repetitions}}
    \]

    \item \textit{args-valid}: Penalizes invalid tool-call arguments.
    \vspace{-0.2em}
    \[
    s_{\text{args-valid}} =
    \begin{cases}
        -0.1, & \text{if arguments are invalid} \\
        0, & \text{otherwise}
    \end{cases}
    \]
\end{compactitem}

\noindent
We set the values for the step rewards such that the accumulated step-level reward for a trajectory with 10 steps would be comparable to the accuracy reward.

\begin{table}[t!]
  \centering
  \fontsize{10}{12}\selectfont
  \resizebox{1.\linewidth}{!}{
  \begin{tabular}{l c | l c}
\toprule
\textbf{Overall} & \textbf{Count} & \textbf{Modality} & \textbf{Count} \\
\midrule
\# samples & 1744 & visual only & 1216 \\
\hspace{1em}-- \# mcq & 802 & verbal only & 134 \\
\hspace{1em}-- \# open-ended & 942 & visual + verbal \textit{(both)} & 394 \\
\midrule
\multicolumn{4}{l}{\textbf{Duration (\textit{avg}: 727 sec.)}} \\
\midrule
\textbf{Bucket (sec.)} & \textbf{Count} & \textbf{Bucket (sec.)} & \textbf{Count} \\
\midrule
0--60        & 261 & 600--1200     & 484 \\
60--180      & 390 & 1200--2400    & 147 \\
180--300     & 116 & 2400+         & 160 \\
300--600     & 186 &               &      \\
\bottomrule
\end{tabular}
}
  \vspaceundertab
  \caption{\textbf{\benchmark Statistics.} Our evaluation set holds 1744 manually verified samples spanning diverse durations, with an emphasis on questions that require visual information to answer.}
    \label{tab:sage_bench}
    \vspace{-0.4cm}
\end{table}

\vspace{0.1em}
\noindent
\textbf{Accuracy Reward.} We compute the outcome reward for a trajectory of length $N$ based on the final answer prediction using an LLM judge (GPT-4o~\cite{openai2024gpt4ocard}) to obtain a binary verdict indicating correctness at the last step.
\vspace{-0.2em}
    \[
    a_N =
    \begin{cases}
        -2.0, & \text{if JSON action string is invalid} \\
        -0.5, & \text{if wrong answer and $N \geq 1$} \\
        +1.25, & \text{if correct answer and visual tools in $\tau_i$} \\
        +1.0, & \text{otherwise}
    \end{cases}
    \]
\noindent
During training and inference, we set $N_{max} = 11$ by default. However, during the RL stage, we find that setting $N_{max} = 6$ for the first 100 steps is necessary for stable training, aligned with findings from a concurrent work for training long-horizon LLM agents~\cite{xi2025agentgymrltrainingllmagents}. Moreover, we penalize the model for predicting a wrong answer with tool calls to compensate for the positive step-level rewards while enforcing the any-horizon nature, i.e., making the model capable of predicting a direct answer. Conversely, we grant a slightly higher reward of 
+1.25 when the answer is correct and \system used visual tools (\texttt{extract-video-parts} or \texttt{ground-event}), reflecting the higher difficulty and importance of getting these tool calls right.

\section{Experiments}
\label{sec:experiments}

\noindent
For our experiments, we finetune for MLLMs, using both cold-start SFT (denoted by \sft) and RL post-training (denoted by \rl) stages to obtain the \model: Molmo-8B, Qwen2.5-VL-7B-Instruct~\cite{Qwen2.5-VL}, Qwen3-VL-4B-Instruct~\cite{qwen3-vl}, and Qwen3-VL-8B-Instruct~\cite{qwen3-vl}. During training, we freeze the visual encoder and projector modules. By default, we use the Qwen3-VL-8B-Instruct as the base \model for all our ablations. We implement the \texttt{transcribe-speech} tool using the Whisper-large-v3~\cite{radford2022whisper} model. We use the Qwen3-VL-30B-A3B-Instruct~\cite{qwen3-vl} model to perform temporal grounding and reasoning with the \texttt{ground-event} and  \texttt{analyze} tools, respectively. 

\subsection{Implementation Details}

\begin{figure}[t!]
\centering
\includegraphics[width=1\linewidth]{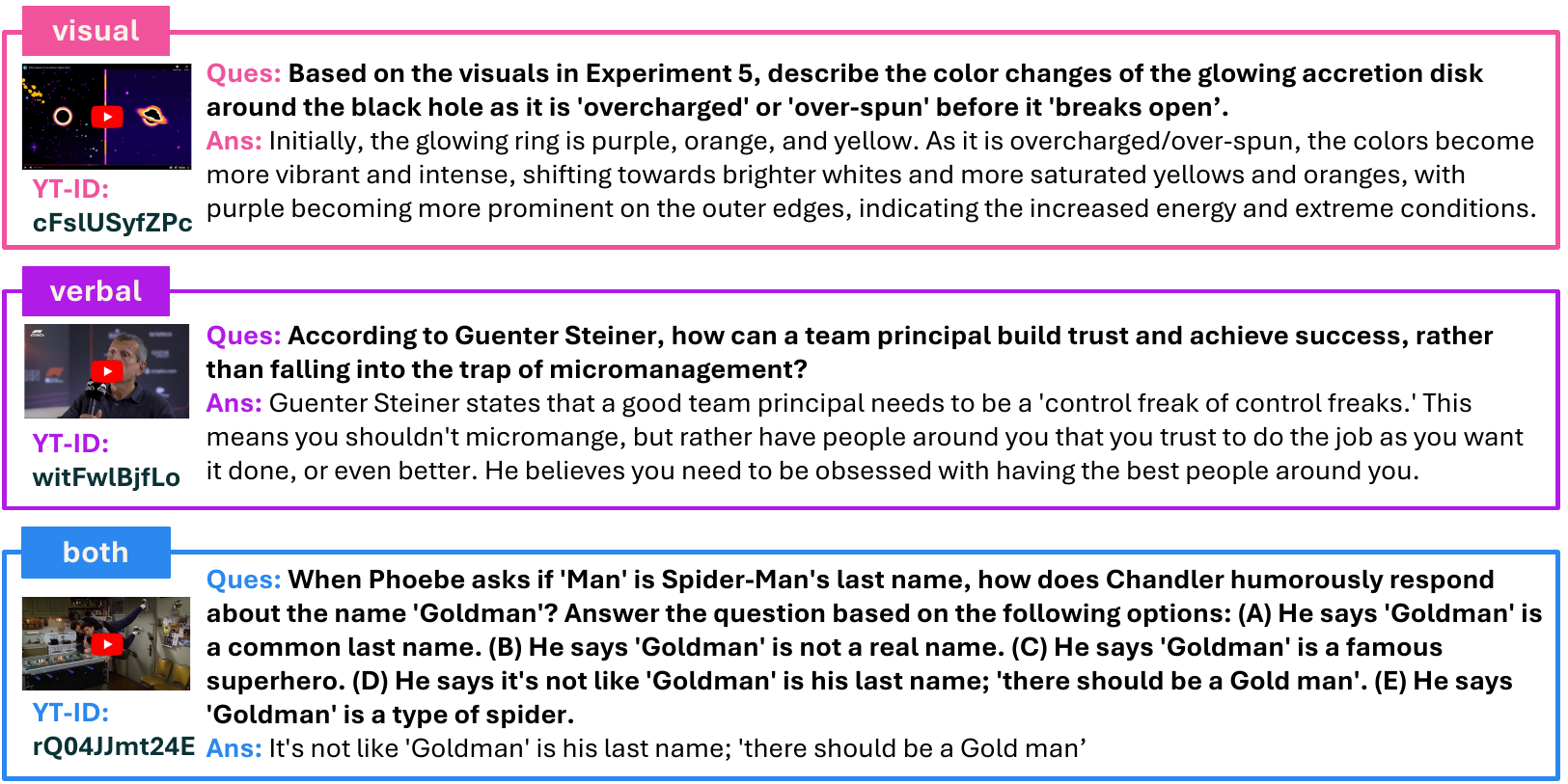} \\
\caption{\textbf{Qualitative Samples from \benchmark.} Our evaluation set contains questions that mirror what a user might naturally ask while or after watching the corresponding video.}
\vspace{-0.4cm}
\label{fig:sage-bench}
\end{figure}

\begin{table*}[t!]
  \centering
  \fontsize{10}{12}\selectfont
  \resizebox{1.\linewidth}{!}{
  \definecolor{darkgray}{gray}{0.90}
\definecolor{lightgray}{gray}{0.95}

\begin{tabular}{ll|cc|c|>{\columncolor{lightgray}}c>{\columncolor{lightgray}}c|
>{\columncolor{lightgray}}c>{\columncolor{lightgray}}c>{\columncolor{lightgray}}c}

\textbf{Method} & \textbf{Orchestrator} &
\multicolumn{2}{c|}{\textbf{Video Reasoning Mode}} &
\textbf{overall} &
\textbf{mcq} &
\textbf{open-ended} &
\textbf{both} &
\textbf{verbal} &
\textbf{visual} \\
\cmidrule(lr){3-4}\cmidrule(lr){5-10}
& & train & eval & (1744) & (802) & (942) & (394) & (134) & (1216) \\
\midrule

{Gemini-2.5-Flash}~\cite{gemini25pushingfrontier} & \na & \direct & \direct & 68.1 &  77.2 & 60.4 & 74.9 & 71.6 & 65.5  \\

\textbf{\system}\textbf{-Flash} (ours) & \textbf{\model:} Gemini-2.5-Flash & \na & \agent & \textbf{71.3} & \textbf{81.2} & \textbf{62.9} & \textbf{76.3} & \textbf{84.3} & \textbf{68.3}  \\
\midrule

{GPT-4o}~\cite{openai2024gpt4ocard} & \na & \direct & \direct & 71.6 & 80.9 & 63.6 & 75.1 & 73.9 & 70.1  \\

\textbf{\system}\textbf{-Flash} (ours) & \textbf{\model:} GPT-4o & \na & \agent & \textbf{73.4} & \textbf{81.0} & \textbf{66.9} & \textbf{78.2} & \textbf{79.9} & \textbf{71.1} \\
\midrule

Video-Thinker-7B~\cite{wang2025video} & \na & \direct & \direct & 41.3 & 70.1 & 16.8 & 48.2 & 41.8 & 39.0 \\

LongVILA-R1-7B~\cite{chen2025longvila-r1} & \na & \direct & \direct & 52.6 & 68.8 & 38.7 & 57.6 & 64.9 & 49.6 \\

VideoRFT-7B~\cite{VideoRFT} & \na & \direct & \direct & 55.3 & 71.6 & 41.4 & 65.2 & 67.2 & 50.7  \\

Video-R1-7B~\cite{video-r1} & \na & \direct & \direct & 57.6 & 73.6 & 43.9 & 67.5 & 67.2 & 53.3  \\

Qwen3-VL-30B-A3B-Instruct~\cite{qwen3-vl} & \na & \direct & \direct & 67.6 & 81.3 & 55.8 & 72.8 & 71.6 & 65.4 \\

VideoAgent~\cite{fan2025videoagent} & GPT-4o & \na & \agent & 42.0 & 52.6 & 32.9 & 42.6	& 29.1 & 43.2 \\

LVAgent~\cite{chen2025lvagent} &
InternVL-8/72B~\cite{chen2023internvl} + LLaVA-Video-72B~\cite{llavavideo} &
\na & \agent & 49.7 & 70.5 & 32.1 & 54.1 & 48.5 & 48.4  \\

LongVT~\cite{yang2025longvt} & LongVT-7B-RFT & \agent & \agent & 46.7 & 68.2 & 28.4 & 51.0 & 37.3 & 46.4 \\

VideoMind~\cite{liu2025videomind} &
VideoMind-7B-Planner & \agent & \agent & 50.0 & 69.7 & 33.2 &  50.8	& 41.8	& 50.7 \\

VideoExplorer~\cite{videoexplorer} &
VideoExplorer-7B-Planner & \agent & \agent & 50.1 & 69.6 & 35.1 & 52.0	& 40.2	& 51.3  \\

VideoChat-R1.5~\cite{yan2025videochatr15} & VideoChat-R1.5-7B-M & \agent & \agent & 54.8 & 73.8 & 38.6 & 55.1 & 48.5 & 55.1 \\

\midrule

Qwen2.5-VL-7B-Instruct~\cite{Qwen2.5-VL} & \na & \direct & \direct & 58.6 &  74.2 & 45.4 & 65.8 & 68.7 & 55.2   \\

\textbf{\system} (ours) & \textbf{\model:} Qwen2.5-VL-7B-Instruct \sftt    &
\agent & \agent & 61.1 & 74.1 & 50.1 & 62.9 & \textbf{69.4} & 59.6  \\

\textbf{\system} (ours) & \textbf{\model:} Qwen2.5-VL-7B-Instruct \sftt \rlt  &
\agent & \agent & \textbf{63.4} & \textbf{77.2} & \textbf{51.5} & \textbf{66.1} & 65.7 & \textbf{62.2}  \\

\midrule

Qwen3-VL-4B-Instruct~\cite{qwen3-vl} & \na & \direct & \direct & 62.7 &  75.8 & 51.6  & 69.3 & 66.4 & 60.2  \\

\textbf{\system} (ours) & \textbf{\model:} Qwen3-VL-4B-Instruct \sftt &
\agent & \agent & 64.6 & 77.3 & 53.7 & 66.2 & 67.2 & 63.7  \\

\textbf{\system} (ours) & \textbf{\model:} Qwen3-VL-4B-Instruct \sftt \rlt &
\agent & \agent & \textbf{68.4} & \textbf{81.3} & \textbf{57.4} & \textbf{78.4} & \textbf{80.6} & \textbf{63.8}   \\
\midrule

Qwen3-VL-8B-Instruct~\cite{qwen3-vl} & \na & \direct & \direct & 64.9 & 77.7 & 54.0 &72.8 & 68.7 & 61.9  \\

\textbf{\system} (ours) & \textbf{\model:} Qwen3-VL-8B-Instruct \sftt  &
\agent & \agent &  63.9 & 77.4 & 52.4 & 72.3 & 74.6 & 60.0   \\

\textbf{\system} (ours) & \textbf{\model:} Qwen3-VL-8B-Instruct \sftt \rlt &
\agent & \agent &  \textbf{68.0} & \textbf{82.6} & \textbf{55.6} & \textbf{75.4} & \textbf{82.8} & \textbf{64.0}   \\

\textbf{\system}\textbf{-Flash} (ours) & \textbf{\model:} Qwen3-VL-8B-Instruct \sftt \rlt &
\agent & \agent & \textbf{71.8} & \textbf{82.8} & \textbf{62.4} & \textbf{75.1} & \textbf{79.1} & \textbf{69.9}    \\

\midrule

Molmo2-8B~\cite{molmo2} & \na & \direct & \direct & 61.8 & 77.9 & 48.1 & 66.8 & 64.2 & 60.0 \\

\textbf{\system} (ours) & \textbf{\model:} Molmo2-8 \sftt  &
\agent & \agent &  63.2 & 75.6 & 52.8 & 67.0 & \textbf{73.9} & 60.9  \\

\textbf{\system} (ours) & \textbf{\model:} Molmo2-8B \sftt \rlt &
\agent & \agent &  \textbf{66.1} & \textbf{78.8} & \textbf{55.2} & \textbf{67.3} & \textbf{73.9} & \textbf{64.8}  \\

\textbf{\system}\textbf{-Flash} (ours) & \textbf{\model:} Molmo2-8B \sftt \rlt &
\agent & \agent &  \textbf{67.8} & \textbf{79.3} & \textbf{58.1} & \textbf{67.5} & \textbf{73.9} & \textbf{67.3}   \\

\bottomrule
\end{tabular}
}
  \vspaceundertab
  \caption{\textbf{Comparison to Baselines.} Using closed-source Gemini-2.5-Flash~\cite{gemini25pushingfrontier} and GPT-4o~\cite{openai2024gpt4ocard} as \model improves upon the base models, showing the effectiveness of our system design. Our trained \model also shows consistent improvements over all the baselines. \system-Flash refers to the setting where we use Gemini-2.5-Flash as the backend model for the \texttt{ground-event} and \texttt{analyze} tools. Existing {\agent} systems exhibit considerably worse performance on open-ended problems compared to our \system.}
    \label{tab:results}
    \vspace{-0.5cm}
\end{table*}

\begin{table}[t!]
  \centering
  \fontsize{10}{12}\selectfont
  \resizebox{1.\linewidth}{!}{
  \begin{tabular}{ll|c|cc}
\toprule
 & \textbf{SAGE-MM} &
\textbf{overall} & \textbf{0--600s} & \textbf{600+s} \\
\cmidrule(lr){2-2} \cmidrule(lr){3-5}
& training & (1473) & (842) & (631) \\
\midrule

Qwen2.5-VL-7B-Instruct~\cite{Qwen2.5-VL} & \na &
32.7 & 37.8 & 25.8 \\

VideoRFT-7B~\cite{VideoRFT} & \na &
30.4 & 33.5 & 26.2 \\

VideoMind-7B~\cite{liu2025videomind} & \na & 30.7 & 34.0	& 26.2 \\

Video-R1-7B~\cite{video-r1} & \na &
31.5 & 36.0 & 25.8 \\

VideoChat-R1.5-7B~\cite{yan2025videochatr15} & \na & \textbf{33.8} & 35.3 & \textbf{31.8} \\

\midrule

\textbf{\system} (ours) & \sft &
28.3 & 30.3 & 24.3 \\

\textbf{\system} (ours) & \sft + \rl &
32.0 & 34.7 & \textbf{28.4} \\

\textbf{\system}\textbf{-Flash} (ours) & \sft + \rl &
\textbf{32.9} & 35.6 & \textbf{29.0} \\

\bottomrule
\end{tabular}









}
  \vspaceundertab
  \caption{\textbf{Performance on MINERVA~\cite{minerva}.} Our \system shows significant improvements on videos longer than 600 seconds.}
  \vspace{-0.3cm}
    \label{tab:ood}
\end{table}

\begin{table}[t!]
  \centering
  \fontsize{10}{12}\selectfont
  \resizebox{1.\linewidth}{!}{
  \begin{tabular}{ll|cc}
\toprule
& \textbf{SAGE-MM} &
\textbf{Video-MMMU} & \textbf{Video-MME} \\
\midrule

Qwen2.5-VL-7B-Instruct~\cite{Qwen2.5-VL} & \na & 57.5 & 63.6 \\
Qwen3-VL-8B-Instruct~\cite{qwen3-vl} & \na & 65.3 & \textbf{66.8} \\
Video-R1-7B~\cite{video-r1} & \na  & 61.5 & 61.2 \\

\midrule

\textbf{\system}\textbf{-Flash} (ours) & Qwen3-VL-8B-Instruct \sftt & \underline{66.9} & 59.4
 \\

\textbf{\system}\textbf{-Flash} (ours) & Qwen3-VL-8B-Instruct \sftt \rlt & \textbf{68.1} & 63.5 \\

\rowcolor{gray!15}
\multicolumn{2}{l|}{\quad\quad - w/o ground-event}  & 65.8 & 65.6 \\
\rowcolor{gray!15}
\multicolumn{2}{l|}{\quad\quad - w/o ground-event \& extract-video-parts}  & 61.8 & \underline{66.2} \\

\bottomrule
\end{tabular}
}
  \vspaceundertab
  \caption{\textbf{Video-MMMU~\cite{hu2025videommmuevaluatingknowledgeacquisition} \& Video-MME~\cite{fu2025videommefirstevercomprehensiveevaluation} (w/o subs).} Our \system-Flash outperforms baselines including Video-R1~\cite{video-r1} on Video-MMMU, demonstrating generalization to knowledge acquisition from videos.}
    \label{tab:eval}
    \vspace{-0.7cm}
\end{table}

\noindent
\textbf{Training Data.} As shown in \cref{tab:data_stats}, we synthesize 99.1k training questions from 6659 videos, covering a wide range of durations. Additionally, we generate 417.7k state–action pairs for \sft. For \rl, we construct a dataset of 7.68k samples, filtered using synthetic tool-call trajectories, where half of the samples required tool calls and the other half had single-turn responses, promoting any-horizon reasoning.

\vspace{0.1em}
\noindent
\textbf{Training Recipe.} During \sft, we train our model for one epoch with a batch size of 64 and an initial learning rate of $1e^ {-5}$ with a linear decay scheduler. We sample 128 frames at 2 FPS and use a temporal pooling factor of 2, setting the maximum and minimum numbers of tokens per frame to 128 and 192, respectively. During \rl, we use a batch size of 16 and rollout eight action trajectories per sample. We use an initial learning rate of $1e^ {-6}$ with a cosine decay scheduler. We set the KL-divergence loss coefficient to 0.005. Note that we report numbers for the model trained for 480 steps during the \rl stage. We train all our models using 16× NVIDIA H100 GPUs during both \sft and \rl.

\vspace{0.1em}
\noindent
\textbf{Evaluation. } We evaluate all {\direct} baselines with 128 sampled frames as input, comparable to \model's input setting. Moreover, we also pass the video transcript as extra context to the {\direct} baselines for fair comparison. For {\agent} baselines, we follow their recommended setup. By default, we use LLM-as-judge (GPT-4o) for evaluating all models on both open-ended and MCQ problems. We set the temperature to 0.0 for all evaluations. However, because the action strings must follow a strict JSON schema, \model occasionally produces malformed outputs. In such cases, we regenerate the response with a temperature of 0.7 for up to four attempts, which may lead to non-deterministic behavior during inference. We serve all supported models using vLLM~\cite{kwon2023efficient} during evaluation.

We share more details, including the system, data generation, and evaluation prompts, in the appendix.

\subsection{\benchmark}

\begin{table}[t!]
  \centering
  \fontsize{10}{12}\selectfont
  \resizebox{1.\linewidth}{!}{
  \begin{tabular}{lc|c|cc|c}

\toprule
train strategy & train mode & eval mode & mcq & open-ended & overall \\ 

\midrule

\multicolumn{2}{l|}{Qwen3-VL-4B-Instruct} & \direct & 75.8 & 51.5 & 62.7 \\
\multicolumn{2}{l|}{Qwen3-VL-4B-Thinking} & \direct & 75.3 & 48.6 & 60.1 \\

\midrule

\sft & \direct & \direct & \textbf{83.2} & 51.1 & 65.8 \\
\sft + \rl & \direct & \direct & 83.0 & 52.0 & 66.3 \\

\midrule

\sft (ours) & \agent & \agent & 77.3 & 53.7 & 64.6 \\

\sft + \rl (ours) & \agent & \agent &  81.3 & \textbf{57.4} & \textbf{68.4} \\

\bottomrule
\end{tabular}
}
  \vspaceundertab
  \caption{\textbf{Training Mode.} Our {\agent} system performs better than the {\direct} baseline, with \rl playing a critical role in the former's success, specifically on open-ended problems.}
  \vspace{-0.5cm}
    \label{tab:abl_train_mode}
\end{table}

\begin{table*}[t!]
  \centering
  \fontsize{10}{12}\selectfont
  \resizebox{1.\linewidth}{!}{
  \begin{tabular}{llc|ccccccc|c}
\toprule
\textbf{Method} & \textbf{Model} & \textbf{Eval Mode} & \textbf{0-60} & \textbf{60-180} & \textbf{180-300} & \textbf{300-600} & \textbf{600-1200} & \textbf{1200-2400} & \textbf{2400+} & {\textbf{overall}} \\ 
\midrule

& & & (261) & (390) & (116) & (186) & (484) & (147) & (160) & (1744) \\
\midrule
 
Qwen3-VL (baseline) & Qwen3-VL-8B-Instruct & \direct & 73.9 & 72.3 & \textbf{81.9} & 71.5 & 55.0 & 59.2 & 47.5 & 64.9 \\



\midrule

\textbf{\system} (ours) & Qwen3-VL-8B-Instruct \sftt & \agent & 74.3 & 68.1 & 75.0 & 72.0 & 56.8 & 55.8 & 48.1 & 63.9 \\

\textbf{\system} (ours) & \textbf{\model}: Qwen3-VL-8B-Instruct \sftt \rlt & \agent & 
\textbf{78.5}~{\scriptsize\textcolor{blue}{\textbf{(+4.6)}}} & 
70.3~{\scriptsize\textcolor{red}{\textbf{(-2.0)}}} & 
77.4~{\scriptsize\textcolor{red}{\textbf{(-4.5)}}} & 
\textbf{72.6}~{\scriptsize\textcolor{blue}{\textbf{(+1.1)}}} & 
\textbf{63.2}~{\scriptsize\textcolor{blue}{\textbf{(+8.2)}}} & 
\textbf{61.9}~{\scriptsize\textcolor{blue}{\textbf{(+2.7)}}} & 
\textbf{53.8}~{\scriptsize\textcolor{blue}{\textbf{(+6.3)}}} & 
\textbf{68.0}~{\scriptsize\textcolor{blue}{\textbf{(+3.1)}}} \\

\textbf{\system}\textbf{-Flash} (ours) & \textbf{\model}: Qwen3-VL-8B-Instruct \sftt \rlt & \agent & 
\textbf{77.8}~{\scriptsize\textcolor{blue}{\textbf{(+3.9)}}} & 
\textbf{73.6}~{\scriptsize\textcolor{blue}{\textbf{(+1.3)}}} & 
80.2~{\scriptsize\textcolor{red}{\textbf{(-1.7)}}} & 
\textbf{76.3}~{\scriptsize\textcolor{blue}{\textbf{(+4.8)}}} & 
\textbf{69.6}~{\scriptsize\textcolor{blue}{\textbf{(+14.6)}}} & 
\textbf{68.0}~{\scriptsize\textcolor{blue}{\textbf{(+8.8)}}} & 
\textbf{56.2}~{\scriptsize\textcolor{blue}{\textbf{(+8.7)}}} & 
\textbf{71.8}~{\scriptsize\textcolor{blue}{\textbf{(+6.9)}}} \\

\bottomrule
\end{tabular}
}
  \vspaceundertab
  \caption{\textbf{Duration-wise Accuracy.} Our \system shows significant improvements on samples belonging to buckets with duration longer than 600 seconds, with even more improvements when using Gemini-2.5-Flash as a tool with \system-Flash.}
    \label{tab:duration}
    \vspace{-0.5cm}
\end{table*}

\noindent
Driven by the limitations of current video reasoning benchmarks due to their purely MCQ nature, we curate our own evaluation set, \textbf{\benchmark}, with a focus on open-ended questions simulating the needs for real-world use-cases for entertainment videos. We begin by sampling a subset of synthetic QnA pairs that is strictly disjoint from the training set (videos can be common) and manually verifying each sample for correctness. Notably, fewer than 5\% of the samples required edits during verification, demonstrating that our synthetic data generation pipeline produces high-quality data at low cost. The statistics of \benchmark are provided in \cref{tab:sage_bench}. We also provide qualitative examples in \cref{fig:sage-bench}.

\subsection{Main Results}

\noindent
In \cref{tab:results}, we compare our \system to {\direct} video reasoning methods, including models trained without RL post-training, like Qwen3-VL-4/8B-Instruct~\cite{qwen3-vl}, and RL-tuned models, like Video-R1~\cite{video-r1}. We also evaluate {\agent} systems like VideoMind~\cite{liu2025videomind} and VideoExplorer~\cite{videoexplorer}.

\vspace{0.1em}
\noindent
\textbf{Effective System Design.} We separately evaluate the performance of our system with two API-based models \model: Gemini-2.5-Flash~\cite{gemini25pushingfrontier} and GPT-4o~\cite{openai2024gpt4ocard}. For this setting, we use Gemini-2.5-Flash as the backend model for the \texttt{ground-event} and \texttt{analyze} tools; therefore, we denote the system as \textbf{\system}\textbf{-Flash}. We observe improvements of up to \textbf{3.2\%} over the base API models, validating the effectiveness of our system design.

\begin{table}[t!]
  \centering
  \fontsize{10}{12}\selectfont
  \resizebox{1.\linewidth}{!}{
  \begin{tabular}{ll|cc|cc|c}
\toprule
\textbf{system} & \textbf{\model} &
\multicolumn{2}{c|}{\textbf{single-turn}} &
\multicolumn{2}{c}{\textbf{multi-turn}} & \textbf{overall} \\
\cmidrule(r{4pt}){2-4} \cmidrule(l){5-7}
\multicolumn{2}{r}{Qwen3-VL-8B-Instruct \textit{(base)}} & {count} & {acc.} & {count} & {acc.} & {acc.} \\
\midrule

\gr{\system-Flash} & \gr{Gemini-2.5-Flash (expert)} & \gr{859} & \gr{76.9} & \gr{885} & \gr{66.0} & \gr{71.3}  \\

\midrule

\textbf{\system} & \sftt (ours) & 706 & 79.0 & 1038 & 53.7 & 64.6 \\

\textbf{\system} & \sftt \rlt (ours) & 948 & 79.6 & 796 & 54.3 & 68.0 \\

\textbf{\system-Flash} & \sftt \rlt (ours) & 940 & 78.8 & 804 & 63.4 & 71.8 \\

\bottomrule
\end{tabular}
}
  \vspaceundertab
  \caption{\textbf{Any-Horizon Reasoning.} \rl refines the tool's overcalling behavior of the \sft model, resulting in a distribution closer to the expert Gemini-2.5-Flash and thus, improved performance.}
    \label{tab:any_horizon}
    \vspace{-0.5cm}
\end{table}

\vspace{0.1em}
\noindent
\textbf{Effective Training Recipe.} As shown in \cref{tab:results}, our \system with a trained \model achieves notable improvements across different base MLLMs. Specifically, \system surpasses Qwen2.5-VL-7B-Instruct by \textbf{4.8\%} overall, with substantial gains of \textbf{+6.1\%} on open-ended and \textbf{+7.0\%} on visual questions, underscoring the effectiveness of our training strategy. Interestingly, models such as Video-R1~\cite{video-r1}, VideoRFT~\cite{VideoRFT}, and VideoExplorer~\cite{videoexplorer}, despite employing finetuned Qwen2.5-VL-7B-Instruct backbones, underperform relative to the base model, particularly on open-ended questions.
Moreover, as shown in the last row of \cref{tab:results}, \system-Flash further improves upon \system by \textbf{3.8\%}, even outperforming the Gemini-2.5-Flash variant of \model. This indicates that our finetuned \model not only learns to invoke tools effectively but also benefits from more accurate tool outputs.

Additionally, we report results with Qwen2.5-VL-7B-Instruct based \model on MINERVA~\cite{minerva}, a complex video reasoning benchmark that covers domains such as sports, short films, and cooking videos. As shown in \cref{tab:ood}, our \system shows an improvement of \textbf{2.6\%} on long videos (duration $>$600 seconds) compared to the base model while outperforming other reasoning models, validating the effectiveness of our approach for long video reasoning.

\vspace{0.1em}
\noindent
\textbf{Generalization to Other Benchmarks.} We also evaluate on Video-MMMU~\cite{hu2025videommmuevaluatingknowledgeacquisition} and Video-MME~\cite{fu2025videommefirstevercomprehensiveevaluation} in \cref{tab:eval}. Our \system-Flash outperforms baselines on Video-MMMU, demonstrating generalization to knowledge acquisition from videos. On Video-MME, the perception-centric nature of the benchmark means \texttt{ground-event}/\texttt{extract-video-parts} tools can hurt performance; disabling them recovers competitive results.

\begin{table}[t!]
  \centering
  \fontsize{10}{12}\selectfont
  \resizebox{1.\linewidth}{!}{
  \begin{tabular}{l|c|ccc}

\toprule

 & overall & both & verbal & visual  \\
\midrule

\textbf{SAGE} (ours) & \textbf{68.0} & \textbf{75.4} & \textbf{82.8} & 64.0 \\

\midrule

w/o ground-event & 67.3 & 72.3 & 79.9 & \textbf{64.3} \\

w/o web-search/parse-website & 65.5 & 70.1 & 80.6 & 62.4 \\

w/o analyze & 63.4 & 70.6 & 80.6 & {59.1} \\

w/o extract-video-parts & 63.0 & 70.8 & 79.9 & \re{58.6 }\\

w/o transcribe-speech & 62.5 & \re{66.8} & \re{46.3} & 62.9 \\

\bottomrule
\end{tabular}
}
  \vspaceundertab
  \caption{\textbf{Dropping Tools during inference.} All tools are critical to the success of \system as a system, with the extract-video-parts and transcribe-speech being the most important ones for answering the visual and verbal/both questions, respectively, as expected.}
    \label{tab:tools}
    \vspace{-0.6cm}
\end{table}

\subsection{Ablations}

\noindent
\textbf{Training Mode.} In \cref{tab:abl_train_mode}, we finetune a Qwen3-VL-4B-Instruct model on the synthetic QnA pairs with {\direct} answering mode under the same data setting. We observe that our {\agent} training recipe outperforms the direct baseline, underscoring the effectiveness of our approach. Specifically, while training the {\direct} baseline with \sft, we supervise the model with only the correct final answer and not the tool call actions. During \rl, we use only the accuracy reward to train the {\direct} baseline.

\noindent
\textbf{Duration-wise accuracy.} We report duration-wise accuracies on \benchmark in \cref{tab:duration}. Notably, our \system exhibits substantially higher gains on longer videos compared to shorter ones, achieving a remarkable \textbf{8.2\%} improvement in the 600–1200 seconds bucket. Incorporating Gemini-2.5-Flash as a tool (\system-Flash) further boosts this gain to \textbf{14.6\%}, with more than 8\% improvements in the 1200–2400 and 2400+ second buckets as well. 

\begin{table}[t!]
  \centering
  \fontsize{10}{12}\selectfont
  \resizebox{1.\linewidth}{!}{
  \begin{tabular}{lc|c|cc}

\toprule
\textbf{method} & \textbf{mode} & \textbf{\#frames} & \textbf{acc.} & \textbf{runtime} (\textit{sec/sample}) \\

\midrule

\multirow{8}{*}{Qwen3-VL-8B-Instruct} & \multirow{8}{*}{\direct}  & 16 & 55.7 & 0.8 \\
& & 32 & 59.3 & 1.1 \\
& & 64 & 62.3 & 2.3 \\
& & 128 & 64.9 & 3.6 \\
& & 256 & 66.1 & 5.7 \\
& & 512 & 65.9 & 7.8 \\
& & 1024 & 62.5 & 18.3 \\
& & 1536 & 60.8 & 27.5 \\
\midrule

VideoRFT-7B~\cite{VideoRFT} &  \direct & 128 & 55.3 & 7.2\\
Video-R1-7B~\cite{video-r1} & \direct & 128 & 57.6 & 7.3 \\
VideoMind-7B~\cite{liu2025videomind} & \agent & --- & 50.0 & 24.7 \\
LVAagent~\cite{chen2025lvagent} & \agent & --- & 49.7 & 92.9 \\
VideoChat-R1.5-7B~\cite{yan2025videochatr15} & \agent & --- & 54.8 & 132.1 \\
VideoExplorer-7B~\cite{videoexplorer} & \agent & --- & 50.1 & 137.7 \\
VideoAgent~\cite{fan2025videoagent} & \agent & --- & 42.0 & 1445.0 \\
\midrule

\textbf{SAGE} & \agent & --- & \textbf{68.0} & \textbf{8.6} \\

\bottomrule
\end{tabular}
}
  \vspaceundertab
  \caption{\textbf{Eval Runtime.} Our \system shows a good performance-efficiency tradeoff owing to its any-horizon reasoning nature.}
    \label{tab:frames}
    \vspace{-0.5cm}
\end{table}

\vspace{0.1em}
\noindent
\textbf{Any-Horizon Reasoning.} A core aspect of \ system's design is to enable any-horizon reasoning, \textit{i.e.}, it is adept at multi-turn reasoning and also directly outputting an answer in a single step. As shown in \cref{tab:any_horizon}, our \sft model, distilled from the expert Gemini-2.5-Flash, inherits strong single-turn ability but tends to show signs of overcalling tools. Incorporating \texttt{RL} further refines this behavior while improving single-turn and multi-turn accuracies.

\vspace{0.1em}
\noindent
\textbf{Importance of Supported Tools.} We ablate the contribution of each tool in \cref{tab:tools}. Dropping the \textit{transcribe-speech}, \textit{extract-video-parts}, and \textit{analyze} tools leads to the most significant performance decline, highlighting their fundamental role in long-video reasoning. In contrast, removing the \textit{ground-event} tool results in only a minor drop, likely due to the tool's inherent inaccuracy. This observation underscores the need for developing better temporal grounding modules.

\vspace{-0.4cm}
\paragraph{Eval Runtime.} In \cref{tab:frames}, we compare the accuracy score and inference runtime per sample for our \system to other existing {\direct}~\cite{video-r1, VideoRFT} and {\agent}~\cite{videoexplorer, fan2025videoagent, chen2025lvagent} baselines and various frame-input setups of the baseline Qwen3-VL-8B-Instruct~\cite{qwen3-vl}. We observe that although the runtime of our \system is comparable to using 512 frames as inputs to Qwen3-VL-8B-Instruct, it shows far superior performance, while only being slower by about 1 second compared to other thinking {\direct} baselines. Moreover, our system is almost 3 times quicker than VideoMind~\cite{liu2025videomind}, the quickest {\agent} baseline, demonstrating the superiority of our system design and training recipe for practical applications over existing systems.

The lower runtime of our framework compared to the {\agent} baselines is primarily due to the baselines' system design, which involves heavy video preprocessing and excessive recurrent model calls. Notably, VideoAgent~\cite{fan2025videoagent} is slowed by a mandatory preprocessing phase in which every 2-second subclip undergoes multi-model analysis for metadata extraction, making it super slow for long videos. Similarly, VideoExplorer~\cite{videoexplorer} suffers from both an initial ~30-second preprocessing delay, arising from dividing the video into multiple subclips for embedding computation, and an inference process involving multiple retrieval steps. Finally, VideoMind~\cite{liu2025videomind} inherently requires more model invocations. This increase can be traced to the system design, which requires repetitive invocations of the verifier module. The verifier is executed multiple times, once for each of the top five potential segments generated by the preceding grounder module, which slows the system.

\section{Conclusion}

In this work, we introduced \system, an any-horizon reasoning system for long video reasoning. We also designed a cost-effective synthetic data generation pipeline for training and evaluating with the target use case of aiding users with open-ended queries while they watch entertainment videos in mind. Through extensive experiments, we validated the effectiveness of our system design and RL post-training recipe at enabling any-horizon reasoning, with considerable gains on videos longer than 10 minutes. We hope our work can serve as a vital proof-of-concept toward training practical {\agent} systems for long video reasoning in the future, moving away from purely {\direct} approaches.

\vspace{0.2cm}
\noindent
\textbf{Future Work.} Looking ahead, training on data from broader domains to handle more use cases is a natural advancement. In addition, integrating more advanced agent-centric policy optimization algorithms~\cite{feng2025group, dong2025aepo, gspo} for \rl presents a promising avenue. Finally, empowering the system to select the appropriate tools and synthesize new ones when necessary~\cite{prabhu2025waltwebagentslearn, vadar} represents an exciting direction.

\vspace{0.2cm}
\noindent
\textbf{Acknowledgements.}This work was in part supported by NSF CAREER Award \#2239840, and the National AI Institute for
Exceptional Education (Award \#2229873) by the National Science Foundation and the Institute of Education Sciences, U.S. Department of Education. We also thank the ML Center @Georgia Tech and PRIOR @Allen AI for supporting this work.

\clearpage
\bibliographystyle{ieee_fullname}
\bibliography{main}

\maketitleappendix

\noindent
In this appendix, we first present additional ablations in \cref{sec:app_abl}, including the effect of video input on \benchmark, the importance of the cold-start SFT stage, and the impact of varying $N_{max}$ during evaluation. {Secondly, we provide qualitative examples from our \benchmark in \cref{sec:app_sage_bench} along with a comparison to existing benchmarks}.
Next, we list all the system prompts used in our work in \cref{sec:prompts}. Lastly, we list qualitative examples from the QnA pair generation pipeline in \cref{sec:qna_data}. Unless mentioned otherwise, we use the Qwen3-VL-8B-Instruct-based \model for all experiments in this appendix and report results after \rl on \benchmark.












\section{Additional Ablations}
\label{sec:app_abl}

\begin{table}[t!]
  \centering
  \fontsize{10}{12}\selectfont
  \resizebox{1.\linewidth}{!}{
  \begin{tabular}{lc|c|ccc}

method & video & overall & both & verbal & visual  \\
\midrule

Qwen3-VL-8B-Instruct & \checkmark & \textbf{64.9} & \textbf{72.8} & 68.7 & \textbf{61.9} \\
Qwen3-VL-8B-Instruct & \xmark & 42.1 & 58.6 & 70.9 & 33.6  \\
\midrule

\textbf{\system} (ours) & \checkmark & \textbf{68.0} & \textbf{75.4} & \textbf{82.8} & \textbf{64.0} \\
\textbf{\system} (ours) & \xmark & 41.0 & 48.0 & 35.8 & 39.3 \\

\bottomrule
\end{tabular}
}
  \vspaceundertab
  \caption{\textbf{Importance of the Video Input.} Access to the video is critical for good performance on \benchmark.}
    \label{tab:memorization}
\end{table}

\paragraph{Importance of the Video Input.} Although we build \benchmark using questions strictly disjoint from the training set, some videos in \benchmark overlap with those seen during training. Therefore, it is essential to assess potential memorization. A natural test is to evaluate whether the model produces correct answers without access to the video. We find no evidence of memorization: performance drops by 27\%, similar to the drop observed for the base Qwen3-VL-8B-Instruct model, as shown in \cref{tab:memorization}, underscoring the validity of our approach and findings.

\begin{table}[t!]
  \centering
  \fontsize{10}{12}\selectfont
  \resizebox{1.\linewidth}{!}{
\begin{tabular}{llc|c|cc}

\model & system & eval mode & overall & mcq & open-ended  \\
\midrule

\multicolumn{6}{l}{\textit{Qwen3-VL-8B-Instruct}} \\
\midrule

\multirow{3}{*}{\sftt} & Qwen3-VL & \direct & 63.6 & 78.4 & 50.9 \\
 & \system & \agent & 63.9 & 77.4 & 52.4 \\
  & \system-Flash & \agent & \textbf{70.5} & 81.0 & 61.5 \\

 \midrule

\multirow{3}{*}{\sftt \rlt} & Qwen3-VL & \direct & 69.8 & 84.0 & 57.6  \\
  & \system & \agent & 68.0 & 82.6 & 55.6 \\
  & \system-Flash & \agent &  \textbf{71.8} & 82.8 & 62.4 \\

\midrule

\multicolumn{6}{l}{\textit{Molmo2-8B}} \\
\midrule

\multirow{3}{*}{\sftt} & Molmo2 & \direct & 55.7 & 68.5 & 44.9 \\
 & \system & \agent & 63.3 & 75.6 & 52.8 \\
  & \system-Flash & \agent & \textbf{69.8} & 79.9 & 61.3 \\

 \midrule

\multirow{3}{*}{\sftt \rlt} & Molmo2 & \direct & 61.0 & 71.7 & 51.9  \\
  & \system & \agent & 66.1 & 78.8 & 55.2 \\
  & \system-Flash & \agent & \textbf{67.8} & 79.3 & 58.1 \\

\bottomrule
\end{tabular}
}
  \vspaceundertab
  \caption{\textbf{Eval Mode.} We find that for the trained \model, {\agent} mode during inference outperforms the {\direct} mode. All the models are trained under the {\agent} paradigm}
  \vspace{-0.2cm}
    \label{tab:eval_mode}
\end{table}

\vspace{-0.4cm}
\paragraph{Eval Mode.} In \cref{tab:eval_mode}, we analyze the effect of eval mode with the trained Qwen3-VL-8B-Instruct \model during inference. We find that the {\agent} mode performs better than {\direct} mode. Surprisingly, the Qwen3-VL~\cite{qwen3-vl} based model shows much better performance than the Molmo2~\cite{molmo2} one with {\direct} eval mode which could be attributed to the two model families' different abilities to learn information directly since all the models are trained under the {\agent} paradigm.

\begin{table*}[t!]
  \centering
  \fontsize{10}{12}\selectfont
  \resizebox{0.7\linewidth}{!}{
  \begin{tabular}{l|cc|cc|c}
\toprule

\textbf{\model} &
\multicolumn{2}{c|}{\textbf{single-turn}} &
\multicolumn{2}{c}{\textbf{multi-turn}} & \textbf{overall} \\
\cmidrule(r{4pt}){2-6}
 & {count} & {acc.} & {count} & {acc.} & {acc.} \\
\midrule

Qwen3-VL-4B-Instruct &  \re{1345} & 54.6 & 399 & 52.8 & \re{54.5} \\

Qwen3-VL-4B-Instruct \sftt \textbf{(ours)} & 691 & 79.5 & 1045 & 54.7 & 64.6 \\

Qwen3-VL-4B-Instruct \sftt \rlt \textbf{(ours)} & 832 & 80.5 & 912 & 57.3 & \textbf{68.4} \\

\midrule

Qwen3-VL-8B-Instruct & 802 & 79.5 & 942 & 53.7 & 63.2\\

Qwen3-VL-8B-Instruct \rlt & \re{1727} & 56.9 & 17 & 23.6 & 56.6 \\

\midrule

Qwen3-VL-8B-Instruct \sftt \textbf{(ours)} & 706 & 79.0 & 1038 & 53.7 & 63.9 \\

Qwen3-VL-8B-Instruct \sftt \rlt \textbf{(ours)} & 948 & 79.6 & 796 & 54.3 & \textbf{68.0} \\

\bottomrule
\end{tabular}
}
  \vspaceundertab
  \caption{\textbf{Importance of \sft.} The cold-start SFT stage is necessary to incentivize multi-turn reasoning during \rl.}
    \label{tab:base_rl}
\end{table*}

\begin{table*}[t!]
  \centering
  \fontsize{10}{12}\selectfont
  \resizebox{1.\linewidth}{!}{
  \begin{tabular}{l|ccccccc} 
\toprule 
\textbf{\model} & \textbf{0-60} & \textbf{60-180} & \textbf{180-300} & \textbf{300-600} & \textbf{600-1200} & \textbf{1200-2400} & \textbf{2400+} \\ 
\midrule 

Qwen3-VL-8B-Instruct \sftt & 2.00 & 2.23 & 2.05 & 2.63 & 3.02 & 3.50 & 3.54 \\ 

Qwen3-VL-8B-Instruct \sftt \rlt & 1.74 & 1.81 & 1.83 & 2.18 & 2.49 & 2.89 & 2.77 \\ 

\bottomrule 
\end{tabular}}
  \vspaceundertab
  \caption{\textbf{\#Turns v/s Video Duration.} The average number of reasoning turns grows gradually with an increase in video duration, demonstrating our \system's any-horizon nature.}
    \label{tab:duration_tool}
\end{table*}

\begin{table*}[t!]
  \centering
  \fontsize{10}{12}\selectfont
  \resizebox{1.0\linewidth}{!}{
  \begin{tabular}{ll|cc|cc|cc|cc|cc|cc|cc}
\toprule

\multicolumn{2}{c|}{$N_{\max} \rightarrow$} & \multicolumn{2}{c|}{\textbf{1}} & \multicolumn{2}{c|}{\textbf{2}} & \multicolumn{2}{c|}{\textbf{3}} & \multicolumn{2}{c|}{\textbf{6}} &
  \multicolumn{2}{c|}{\re{\textbf{11} (default)}} & \multicolumn{2}{c|}{\textbf{13}} & \multicolumn{2}{c}{\textbf{16}} \\

\midrule
system & \model &
acc. & no ans. & acc. & no ans. & acc. & no ans. &
acc. & no ans. & acc. & no ans. & acc. & no ans. & acc. & no ans. \\
\midrule

{\system} & Qwen3-VL-8B-Instruct \sftt 
  & 33.8 & 60.8 & 46.0 & 45.0 & 57.0 & 23.4 & 62.1 & 5.5 & 63.9 & 3.3 &  63.8 & 3.5 & 64.6 & 3.0 \\

{\system} & Qwen3-VL-8B-Instruct \sftt \rlt 
  & 43.4 & 46.8 & 56.7 & 30.4 & 62.8 & 20.2 & 66.6 & 3.9 & 68.0 & 1.3 & 67.8 & 1.1 & 67.9 & 1.1 \\

{\system-Flash} & Qwen3-VL-8B-Instruct \sftt \rlt 
  & 43.5 & 47.1 &  57.1 & 30.1 & 63.2 & 21.2 & 70.7 & 6.3 & 71.8 & 3.3 & 72.2 & 2.9 & 71.9 & 2.9  \\

\bottomrule
\end{tabular}
}
  \vspaceundertab
  \caption{\textbf{Effect of $N_{max}$.} Limiting the total number of turns to 11 is optimal as our \rl recipe enforces the ability to produce an answer in as many turns. \textit{no ans.} denotes the percentage of samples where an answer could not be produced. \textit{acc.} denotes the accuracy score.}
    \label{tab:turns}
\end{table*}


\vspace{-0.4cm}
\paragraph{Importance of \sft.} Our \system is designed so that any MLLM with function-calling capabilities can be used as the \model. In \cref{tab:base_rl}, we evaluate the base Qwen3-VL~\cite{qwen3-vl} models as \model, without any finetuning. We observe that Qwen3-VL-4B-Instruct is not an effective orchestrator: it rarely engages in multi-turn reasoning and attains low accuracy, indicating that \sft is essential before applying \rl.

Interestingly, the base Qwen3-VL-8B-Instruct model behaves differently. It is a noticeably stronger function caller, demonstrating a more reasonable balance between single-turn and multi-turn reasoning. This motivates us to apply \rl directly on top of the base model to assess the importance of \sft for the 8B variant. Surprisingly, \rl without \sft fails, \textit{i.e.}, the model collapses to single-turn reasoning. We hypothesize that this is due to the base model's training objective, which strongly biases it toward directly producing final answers, making \sft necessary to incentivize~\cite{wang2025octothinker} any-horizon reasoning during \rl. While it is possible that a heavily engineered \rl recipe could overcome this, we do not pursue this direction, as \sft is far simpler and cheaper than extensive hyperparameter tuning during \rl.

\vspace{-0.4cm}
\paragraph{\#Turns v/s Video Duration.} In \cref{tab:duration_tool}, we report the average number of reasoning turns across all samples grouped by video duration buckets. We observe a gradual increase in the number of turns as video length increases, indicating that \system naturally adapts its trajectory length to the temporal horizon of the input. Shorter videos lead to shorter reasoning trajectories, whereas longer videos elicit more extended ones, aligned with our design objective of instilling any-horizon reasoning into the system.

\vspace{-0.4cm}
\paragraph{Effect of $N_{max}$.} We study the effect of varying $N_{max}$ during evaluation in \cref{tab:turns}. We find that setting $N_{max} = 11$ achieves high accuracy while keeping the number of unanswered samples low, with only minimal gains from further increases in $N_{max}$. This demonstrates the effectiveness of our \rl recipe in enforcing answer prediction within an 11-step reasoning horizon.

\vspace{-0.4cm}
\paragraph{Variance on \benchmark.} We analyze the variance in performance on \benchmark across five different runs (with temperature of 1.0) in \cref{tab:variance}. We find a low standard deviation of 0.22 for the base Qwen3-VL-8B-Instruct model. The low variance indicates that the performance improvements from our \system are statistically significant.

\begin{table}[t!]
  \centering
  \fontsize{10}{12}\selectfont
  \resizebox{1.\linewidth}{!}{
\begin{tabular}{ccccc|cc}
run-1 & run-2 & run-3 & run-4 & run-5 & mean & std \\
\midrule
64.9 & 64.6 & 64.9 & 65.2 & 65.1 & \textbf{64.9} & \textbf{0.22} \\
\bottomrule
\end{tabular}
}
  \vspaceundertab
  \caption{\textbf{Variance on \benchmark.} We find a low standard deviation of 0.22 across five different runs with Qwen3-VL-8B-Instruct with temperature set to 1.0.}
  \vspace{-0.2cm}
    \label{tab:variance}
\end{table}

\vspace{-0.4cm}
\paragraph{Per-Tool Accuracy.} We report per-tool accuracy in \cref{tab:per-tool}, evaluating performance when only a single tool is available. We observe that \texttt{extract-video-parts}/\texttt{ground-event} perform the worst, attributed to their dependence on tools like \texttt{analyze}/\texttt{extract-video-parts} for local segment processing. The best individual tool performance comes from \texttt{transcribe-speech}, highlighting the importance of speech information for long video understanding.

\begin{table}[t!]
  \centering
  \fontsize{10}{12}\selectfont
  \resizebox{1.\linewidth}{!}{
  \begin{tabular}{lc|lc|lc|lc}

\toprule
 \textbf{all-tools} & \textbf{71.8} & transcribe-speech & 61.1 &  \multicolumn{3}{l}{web-search/parse-website} & 58.6 \\

\midrule
analyze & 58.6 & \re{extract-video-parts} & \re{50.2} & \re{ground-event} & \re{50.3} &  {no-tools} & {53.0} \\

\bottomrule
\end{tabular}
}
  \vspaceundertab
  \caption{\textbf{Per-tool performance comparison on \benchmark.}}
    \label{tab:per-tool}
\end{table}

\vspace{-0.4cm}
\paragraph{Qualitative Failure Analysis.} We observe a few failure patterns of \system. First, \system may overcall tools when a tool invocation fails, resulting in retries that consume reasoning steps. Second, inaccurate temporal grounding can result in missing video segments, leading to incomplete context. Lastly, misleading transcript or web information can result in incorrect answers.

\section{SAGE-Bench}
\label{sec:app_sage_bench}

\begin{figure*}[t!]
\centering
\includegraphics[width=1.\linewidth]{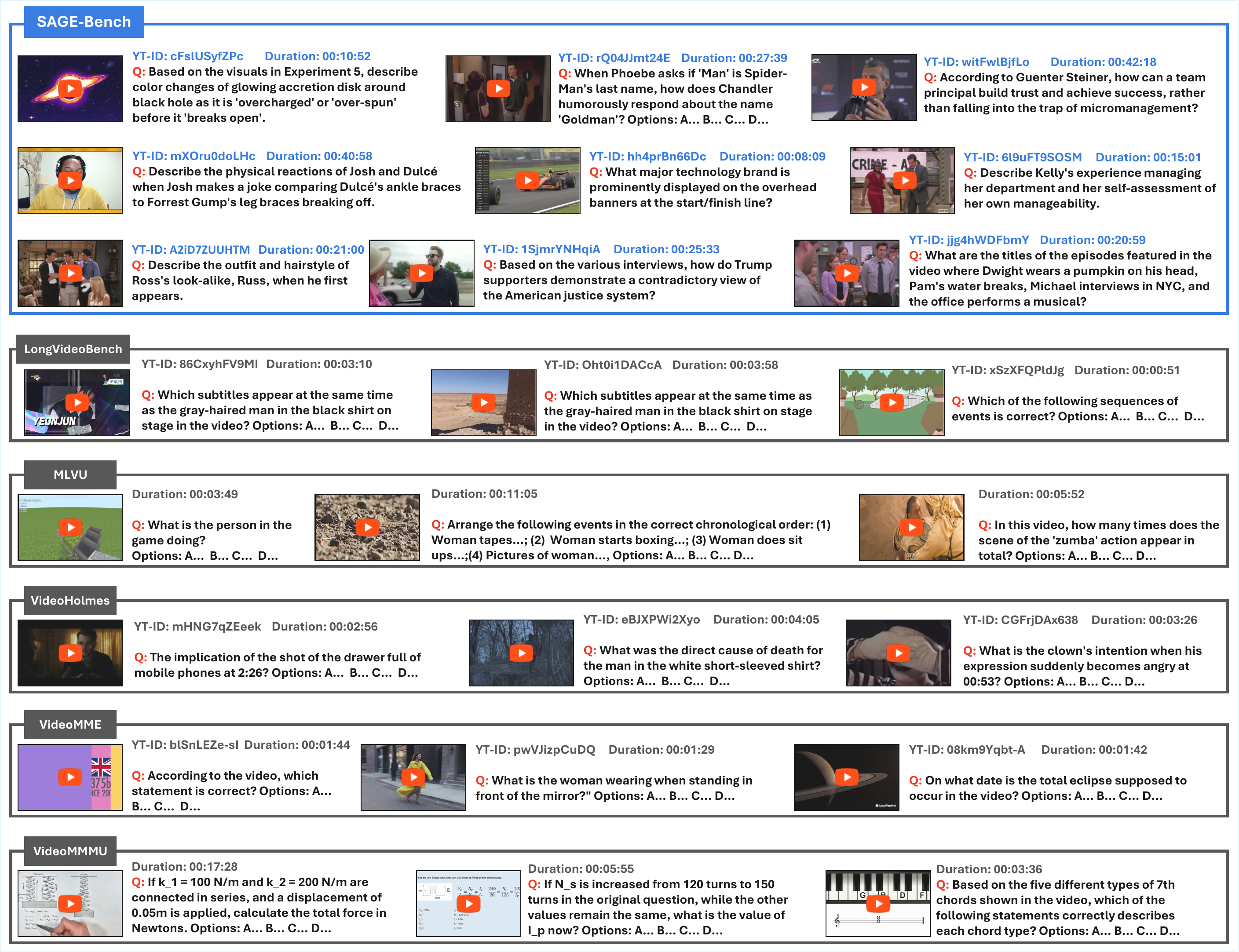} \\
\vspace{-0.1cm}
\caption{\textbf{Comparing \benchmark to existing benchmarks.}  \benchmark contains samples covering both practical scenarios (IDs: \textit{witFwlBjfLo, jjg4hWDFbmY, hh4prBn66Dc, 1SjmrYNHqiA}) and diagnostic cases. Representative examples for other benchmarks are sourced from their respective websites or papers.}
\label{fig:sage-app}
\end{figure*}

Compared to existing video understanding benchmarks, \benchmark demonstrates two distinct advantages:


\vspace{-0.4cm}
\paragraph{High-Quality Open-Ended Questions.} 
As illustrated in Fig.~\ref{fig:sage-app}, existing popular benchmarks~\cite{zhou2025mlvubenchmarkingmultitasklong, cheng2025video, hu2025videommmuevaluatingknowledgeacquisition, wu2024longvideobenchbenchmarklongcontextinterleaved, fu2025videommefirstevercomprehensiveevaluation} rely purely on multiple-choice questions (MCQs). In contrast, \benchmark utilizes open-ended questions with an unbounded answer space, aligning more closely with practical user situations. 

\vspace{-0.4cm}
\paragraph{Dual Focus on Diagnostic and Practical Evaluation.} 
While AI systems are ultimately intended for real-world deployment, existing benchmarks often include diagnostic questions to gauge models' visual understanding, such as temporal ordering tasks in MLVU~\cite{zhou2025mlvubenchmarkingmultitasklong}. \benchmark incorporates both diagnostic questions to test fundamental model capabilities and practical questions that users may have while watching entertainment videos, ensuring that our benchmark evaluates not only technical proficiency but also the model's utility in practical scenarios.

\section{System Prompts}
\label{sec:prompts}

We provide information about the system prompts used for different purposes in this work below:

\begin{compactitem}
    \item {QnA Pair Generation}: \cref{fig:ques_prompt}.
    \item {LLM-Judge Evaluation}: \cref{fig:judge_prompt}.
    \item {\system Stage-1 (Context VLM)}: \cref{fig:cvlm_prompt}.
    \item {\system Stage-2 (Iterative Reasoner)}: \cref{fig:ir_prompt}.
    \item {\texttt{ground-event} tool}: \cref{fig:ground_prompt}.
    \item {Reasonable Tool Step Reward Computation}: \cref{fig:rt_prompt}.
    \item {{\direct} baselines Evaluation}: \cref{fig:baseline_prompt}.
    
\end{compactitem}

\lstdefinelanguage{json}{
    basicstyle=\small\ttfamily,
    stepnumber=1,
    numbersep=4pt,
    showstringspaces=false,
    breaklines=true,
    breakatwhitespace=true,
    commentstyle=\color{green!60!black}\rmfamily,
    morecomment=[l]{//},
    morecomment=[l]{\#},
    literate=
     *{true}{{{\color{blue}true}}}{4}
      {false}{{{\color{blue}false}}}{5}
      {null}{{{\color{blue}null}}}{4},
    alsoletter={:,-,_},
    stringstyle=\color{red!70!black},
    morestring=[b]",
}

\begin{figure*}[t]
\centering
\begin{tcolorbox}[
    title={System Prompt to \underline{generate QnA pairs} using \underline{Gemini-2.5-Flash}},
    colback=white,
    colframe=black!70,
    width=\textwidth,
    top=2mm,
    bottom=2mm]

{\small
You are a specialized question generator. Your primary function is to generate 10--20 questions based on the provided video which can be upto 2 hours (7200 seconds) long.

\smallskip
\noindent
- Pay attention to what modality information is needed to answer the question. You should generate questions that a viewer may be interested in and require visual, verbal, and or both in a balanced manner.

- You MUST give atleast four questions that cannot be answered with verbal information and require visual information.

- Also, it's okay to give questions that are not answerable from the video but can be answered with a web search.

- Generate a mix of open ended and multiple choice questions which are both hard and easy to answer. Err on the side of hard if you are unsure.

\smallskip
\noindent
The duration of the video is \textcolor{red}{\texttt{<<<video\_duration>>>}} seconds ( \textcolor{red}{\texttt{<<<timestamp\_format>>>}} in HH:MM:SS format). 

First think about the facts from the video and then generate questions about those. The questions could refer to the part of the video that spans across 10 seconds long but most MUST refer to the timeframes atleast a few minutes long.

Your timestamps MUST be in HH:MM:SS format.

\smallskip
\noindent
\textbf{Output Format.} You MUST follow this format and MUST be between the \texttt{<json>} and \texttt{</json>} tags:
}

\begin{lstlisting}[language=json, escapechar=@, basicstyle=\small\ttfamily]
<json>
{
  "timestamp_format":@\textcolor{black}{"HH:MM:SS"}@,
  "num_questions": @\textit{<number of questions generated>}@,
  "questions": [
    {
      "index": @\textit{<index\_of\_question\_out\_of\_total\_question>}@,
      "type": @\textcolor{black}{"type\_of\_question"}@, // can be mcq or open_ended
      "difficulty": @\textit{<difficulty\_of\_question>}@, // can be easy, medium, hard
      "difficulty_rationale": @\textit{<why-this-difficulty>}@,
      "modality": @\textit{<modality\_of\_question>}@, // can be visual, verbal, or both
      "modality_rationale": @\textit{<why-this-modality>}@,
      "answer": @\textit{<answer\_text>}@, // answer for the question, if the type of question is mcq, then this is the text for the correct option, otherwise this is the answer text for the open ended question
      "question": @\textit{<question\_text>}@,
      "options": [ // if the type of question is mcq, then this is a list of options, otherwise this is null
        @\textit{<option\_1>}@, @\textit{<option\_2>}@, @\textit{<option\_3>}@, @\textit{<option\_4>}@, @\textit{<option\_5>}@, @\textit{<option\_6>}@
      ]
      "requires_web_search": @\textit{<true | false>}@, // if the question requires a web search to be answered, then this is true, otherwise this is false
      "why_web_search": @\textit{<reasoning for why web search is needed to answer the question>}@, // if the question requires a web search to be answered, then this is the reasoning for why web search is needed to answer the question, otherwise this is null
      "final_timestamp": @\textit{<duration\_of\_the\_video>}@, # HH:MM:SS
      "start_timestamp": @\textit{<start\_timestamp\_of\_question>}@, # HH:MM:SS
      "end_timestamp": @\textit{<end\_timestamp\_of\_question>}@, # HH:MM:SS
      "compute_percent_video_parsed": @\textit{<think carefully and predict accurate percent\_video\_parsed, show calculation here>}@,
      "percent_video_parsed": @\textit{<percentage\_of\_the\_video\_parsed\_upto\_this\_question>}@ # [(end_timestamp(seconds)/final_timestamp(seconds)) * 100] MUST go upto atleast 90 if not 100 for atleast one question
    },
    ...
  ]
}
</json>
\end{lstlisting}

{
\noindent
This output will be converted to a JSON dict later on, you MUST use the correct syntax.
}
\end{tcolorbox}

\vspace{-0.4cm}
\caption{System Prompt to \underline{generate QnA pairs} using \underline{Gemini-2.5-Flash}. Placeholder text to be replaced by the corresponding values are in \re{red}.}
\label{fig:ques_prompt}
\end{figure*}

\begin{figure*}[t]
\centering
\begin{tcolorbox}[
    title={System Prompt for the \underline{LLM-Judge} during evaluation and \textbf{{RL}} to compute accuracy},
    colback=white,
    colframe=black!70,
    width=\textwidth,
    top=2mm,
    bottom=2mm]

{
Compare the model prediction and the ground truth and determine if they convey the same meaning for the question:

\smallskip
\noindent
{Question:}  \textcolor{red}{\texttt{\{question\}}}

\smallskip
\noindent
{Model Prediction:}  \textcolor{red}{\texttt{\{hypothesis\}}}

\noindent
{Ground Truth:}  \textcolor{red}{\texttt{\{reference\}}}

\smallskip
\noindent
You MUST respond with the verdict as \texttt{'True'} if they match semantically or \texttt{'False'} if they don't match.

\smallskip
\noindent
Answer in the following format:
}

\begin{lstlisting}[language=python, escapechar=@, basicstyle=\small\ttfamily]
Reasoning: @\textcolor{black}{<Reasoning for the verdict>}@
Verdict: @\textcolor{black}{<True/False>}@
\end{lstlisting}

\end{tcolorbox}

\vspace{-0.4cm}
\caption{System Prompt for the \underline{LLM-Judge} during evaluation and \rl to compute accuracy. Placeholder text to be replaced by the corresponding values are in \re{red}.}
\label{fig:judge_prompt}
\end{figure*}

\lstdefinelanguage{json}{
    basicstyle=\small\ttfamily,
    stepnumber=1,
    numbersep=4pt,
    showstringspaces=false,
    breaklines=true,
    breakatwhitespace=true,
    commentstyle=\color{green!60!black}\rmfamily,
    morecomment=[l]{//},
    morecomment=[l]{\#},
    literate=
     *{true}{{{\color{blue}true}}}{4}
      {false}{{{\color{blue}false}}}{5}
      {null}{{{\color{blue}null}}}{4},
    alsoletter={:,-,_},
    stringstyle=\color{red!70!black},
    morestring=[b]",
}

\begin{figure*}[t]
\centering
\begin{tcolorbox}[
    title={\system Stage-1: \underline{Context VLM} System Prompt},
    colback=white,
    colframe=black!70,
    width=\textwidth,
    top=2mm,
    bottom=2mm]

{
You are a specialized Context VLM (Video Language Model) designed to analyze video content and determine the appropriate context for further processing. Your primary functions are to:

\smallskip
\noindent
- Analyze the given video and query

- Recommend the next appropriate tool or sequence of tools

- Suggest specific arguments to pass to those tools

\smallskip
\noindent
Your output MUST follow this structure and MUST be between the \texttt{<json>} and \texttt{</json>} tags:
}

\begin{lstlisting}[language=json, escapechar=@, basicstyle=\small\ttfamily]
<json>
{
  "video_context": @\textcolor{black}{<visual\_context>}@,
  "query_intent": @\textcolor{black}{<user's\_intent>}@,
  "final_answer": @\textcolor{black}{"Direct and concise answer to the user's query, if and only if the query is answerable based on current context. Otherwise, this should be null."}@,
  "recommended_tools": {
    "needed":true | false,
    "why_no_tool": @\textcolor{black}{"Only if no more tool call is needed"}@,
    "tool_calls": [
      {
        "rationale": @\textcolor{black}{"Why this tool is the best next step"}@,
        "name": @\textcolor{black}{<name\_of\_tool>}@,
        "arguments": {
          "arg1": @\textcolor{black}{<value1>}@,
          "arg2": @\textcolor{black}{<value2>}@
        }
      }
    ]
  }
}
</json>
\end{lstlisting}

{
\noindent
The available tools are:  \textcolor{red}{\texttt{<<<tools>>>}}
}
\end{tcolorbox}
\vspace{-0.4cm}
\caption{\system Stage-1: \underline{Context VLM} System Prompt. Placeholder text to be replaced by the corresponding values are in \re{red}.}
\label{fig:cvlm_prompt}
\end{figure*}

\begin{figure*}[t]
\centering
\begin{tcolorbox}[
    title={\system Stage-2: \underline{Iterative Reasoner} System Prompt},
    colback=white,
    colframe=black!70,
    width=\textwidth,
    top=2mm,
    bottom=2mm]

{
You are a reasoning agent. Your primary goal is to determine whether the available visual context and tool call information contains sufficient information to answer the user's query. If not, recommend which tools to invoke next, with appropriate arguments.

\smallskip
\noindent
Do \textbf{not} make assumptions beyond the evidence provided. Avoid fabricating facts.

\smallskip
\noindent
Output Format. You MUST follow this format and MUST be between the \texttt{<json>} and \texttt{</json>} tags:
}

\begin{lstlisting}[language=json, escapechar=@, basicstyle=\small\ttfamily]
<json>
{
  "answerable": {
    "verdict": true | false,
    "reasoning": @\textcolor{black}{"Why the available information is sufficient or not"}@
  },
  "final_answer": @\textcolor{black}{"If the query is answerable, otherwise null."}@,
  "recommended_tools": {
    "needed": true | false,
    "why_no_tool": @\textcolor{black}{"Only if no more tool call is needed"}@,
    "tool_calls": [
      {
        "rationale": @\textcolor{black}{"Why this tool is the best next step"}@,
        "name": @\textcolor{black}{<name\_of\_tool>}@,
        "arguments": {
          "arg1": @\textcolor{black}{<value1>}@,
          "arg2": @\textcolor{black}{<value2>}@
        }
      }
    ]
  }
}
</json>
\end{lstlisting}

{\small
\noindent
The available tools are:  \textcolor{red}{\texttt{<<<tools>>>}}
}
\end{tcolorbox}
\vspace{-0.4cm}
\caption{\system Stage-2: \underline{Iterative Reasoner} System Prompt. Placeholder text to be replaced by the corresponding values are in \re{red}.}
\label{fig:ir_prompt}
\end{figure*}

\begin{figure*}[t]
\centering
\begin{tcolorbox}[
    title={System Prompt for the \textbf{\texttt{ground-event}} tool},
    colback=white,
    colframe=black!70,
    width=\textwidth,
    top=2mm,
    bottom=2mm]

{
Given the below event, identify the timestamps for the event in the video.

\noindent
You are given the snippet belonging to the period between \re{\texttt{<<<begin>>>}} and \re{\texttt{<<<end>>>}} (in HH:MM:SS format) of the original video.

\noindent
You should set the start and end timestamps in your answer accordingly to align it to the original video.

\noindent
If the event does not occur, set start and end to null.

\smallskip
\noindent
Event:

\noindent
\re{\texttt{<<<event>>>}}

\smallskip
\noindent
Output Format. You MUST follow this format and MUST be between the \texttt{<json>} and \texttt{</json>} tags:
}

\begin{lstlisting}[language=json, escapechar=@, basicstyle=\small\ttfamily]
<json>
{
  "name": @\textcolor{black}{"the name of the event"}@,
  "timestamps": {
    "start": @\textcolor{black}{"start\_time"}@, #HH:MM:SS
    "end": @\textcolor{black}{"end\_time"}@ #HH:MM:SS
  }
}
</json>
\end{lstlisting}

\end{tcolorbox}

\vspace{-0.4cm}
\caption{System Prompt for the \textbf{\texttt{ground-event}} tool. Placeholder text to be replaced by the corresponding values are in \re{red}.}
\label{fig:ground_prompt}
\end{figure*}

\begin{figure*}[t]
\centering
\begin{tcolorbox}[
    title={System Prompt for the reasonable-tool ($s_{\text{reasonable-tool}}$) step reward during \textbf{{RL}}},
    colback=white,
    colframe=black!70,
    width=\textwidth,
    top=2mm,
    bottom=2mm]

{
Below is the reasoning trace for calling a sequence of tools for finding the answer to the question:

\smallskip
\noindent
{Question:} \re{\texttt{\{question\}}}

\smallskip
\noindent
{Reasoning Trace:} \re{\texttt{\{reasoning\_trace\}}}

\smallskip
\noindent
{Predicted Answer:} \re{\texttt{\{predicted\_answer\}}}

\smallskip
\noindent
You MUST respond with the verdict as \texttt{'True'} if the reasoning trace makes sense for the question leading to the predicted answer or \texttt{'False'} if it doesn't.

\noindent
You MUST penalize repetitive tool calls if they are not needed.

\noindent
Answer in the following format:
}

\begin{lstlisting}[language=python, escapechar=@, basicstyle=\small\ttfamily]
Reasoning: @\textcolor{black}{<Reasoning for the verdict>}@
Verdict: @\textcolor{black}{<True/False>}@
\end{lstlisting}

\end{tcolorbox}

\vspace{-0.4cm}
\caption{System Prompt for the reasonable-tool ($s_{\text{reasonable-tool}}$) step reward during \rl. Placeholder text to be replaced by the corresponding values are in \re{red}.}
\label{fig:rt_prompt}
\end{figure*}

\begin{figure*}[t]
\centering
\begin{tcolorbox}[
    title={Prompt for evaluating \textcolor{black}{{\direct}} baselines},
    colback=white,
    colframe=black!70,
    width=\textwidth,
    top=2mm,
    bottom=2mm]

{
You will be given a question about a video. You are provided frames from the video, sampled evenly across the video.

\smallskip
\noindent
Transcript: \re{\texttt{<<<asr\_transcript>>>}}

\smallskip
\noindent
Question: \re{\texttt{<<<question>>>}}

\smallskip
\noindent
Respond to the user's question.
}

\end{tcolorbox}
\vspace{-0.5cm}
\caption{Prompt for evaluating {{\direct}} baselines. Placeholder text to be replaced by the corresponding values are in \re{red}.}
\label{fig:baseline_prompt}
\end{figure*}

\section{Qualitative Examples}
\label{sec:qna_data}

\begin{figure*}[t!]
\centering
\includegraphics[width=1.\linewidth]{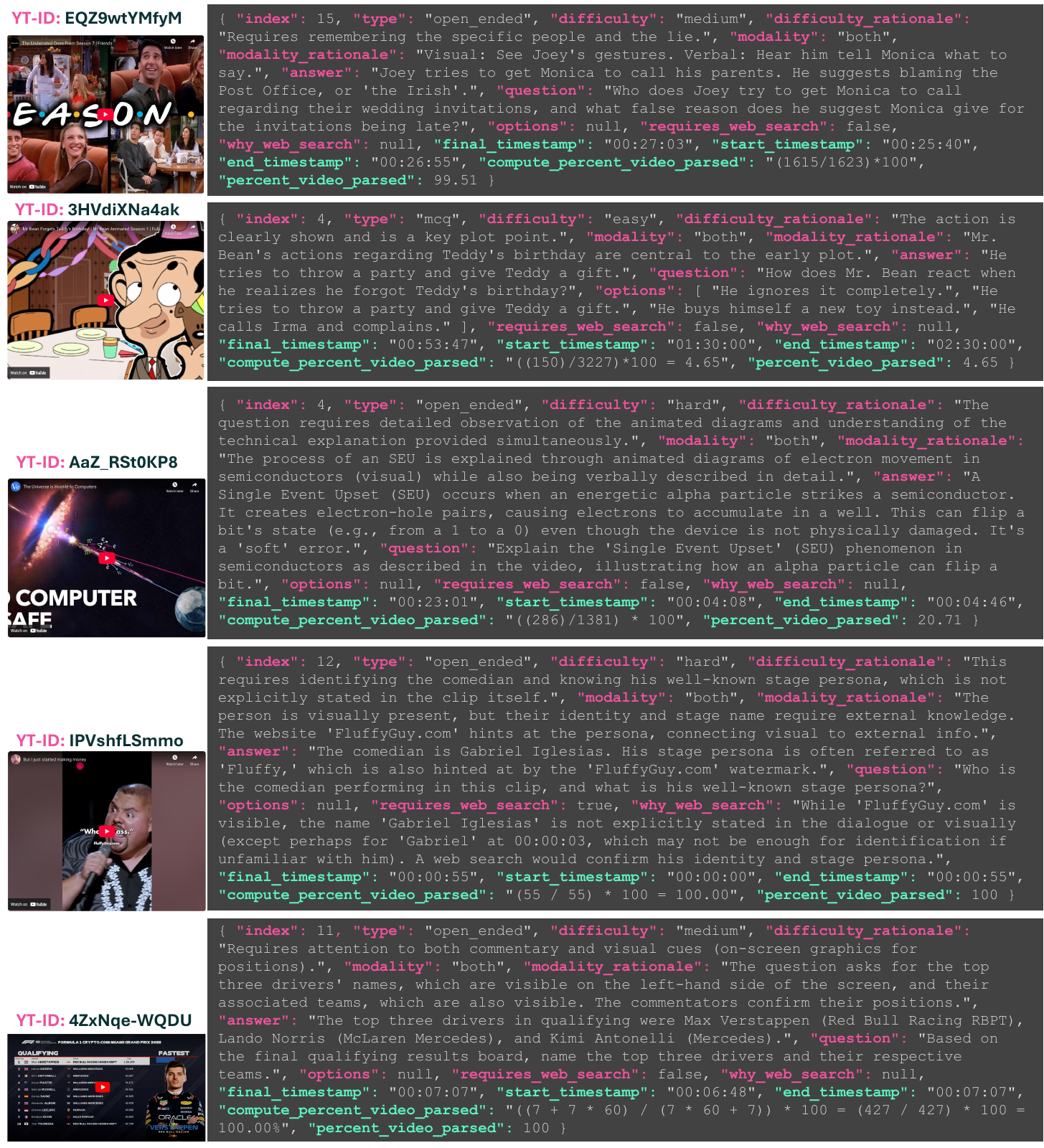} \\
\vspace{-0.1cm}
\caption{\textbf{QnA Pairs Qualitative Samples}. Given our system prompt, Gemini-2.5-Flash can generate high-quality QnA pairs of varying difficulty levels and types (open-ended and MCQ) that cover the entire video.}
\label{fig:ques2}
\end{figure*}

\begin{figure*}[t!]
\centering
\includegraphics[width=1.\linewidth]{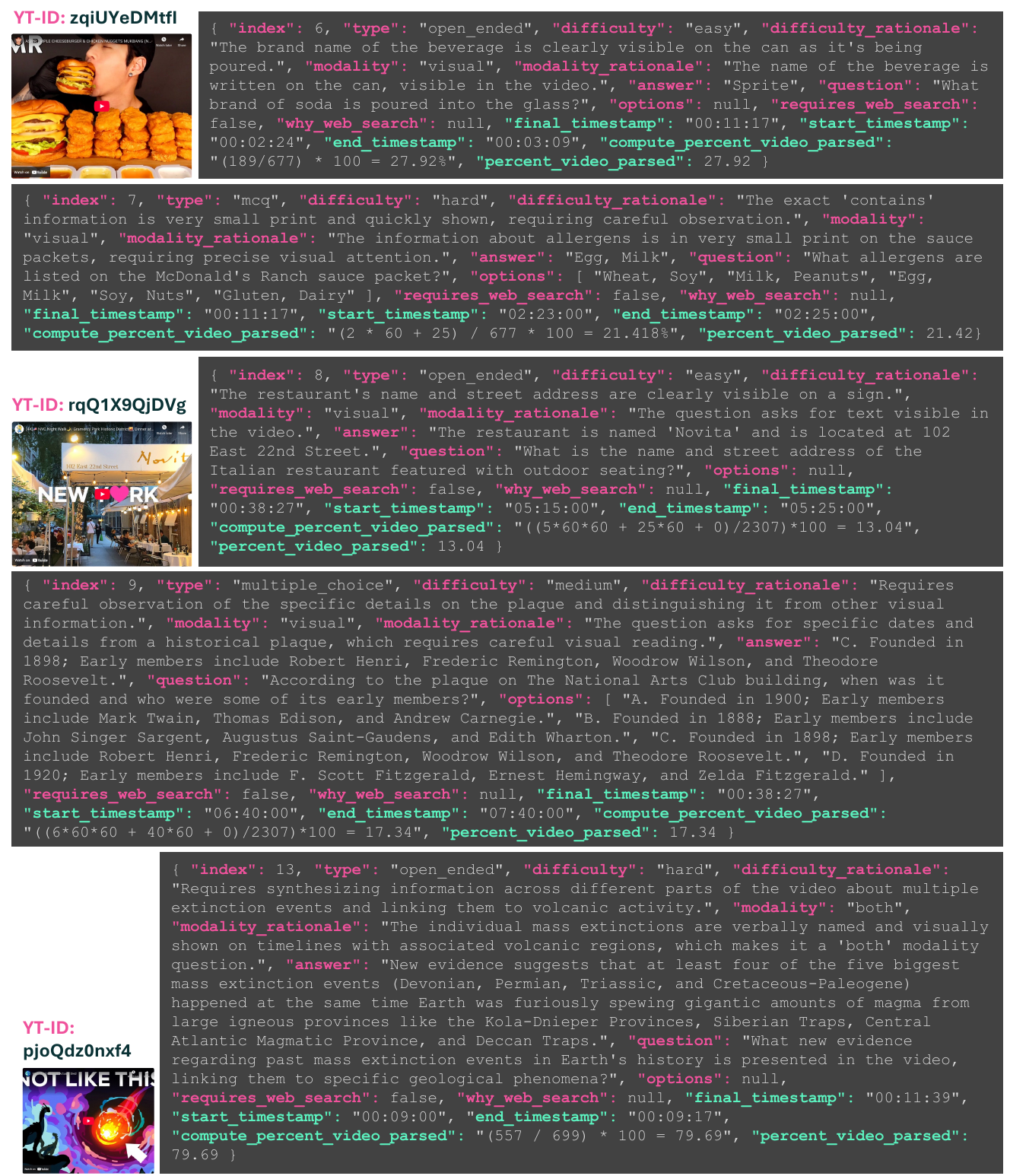} \\
\vspace{-0.1cm}
\caption{\textbf{QnA Pairs Qualitative Samples}. Given our system prompt, Gemini-2.5-Flash can generate high-quality QnA pairs of varying difficulty levels and types (open-ended and MCQ) that cover the entire video.}
\label{fig:ques}
\end{figure*}

\paragraph{QnA Pairs.} We display some samples of the generated QnA pairs in \cref{fig:ques} and \cref{fig:ques2}.

\begin{figure*}[t!]
\centering
\includegraphics[width=0.75\linewidth]{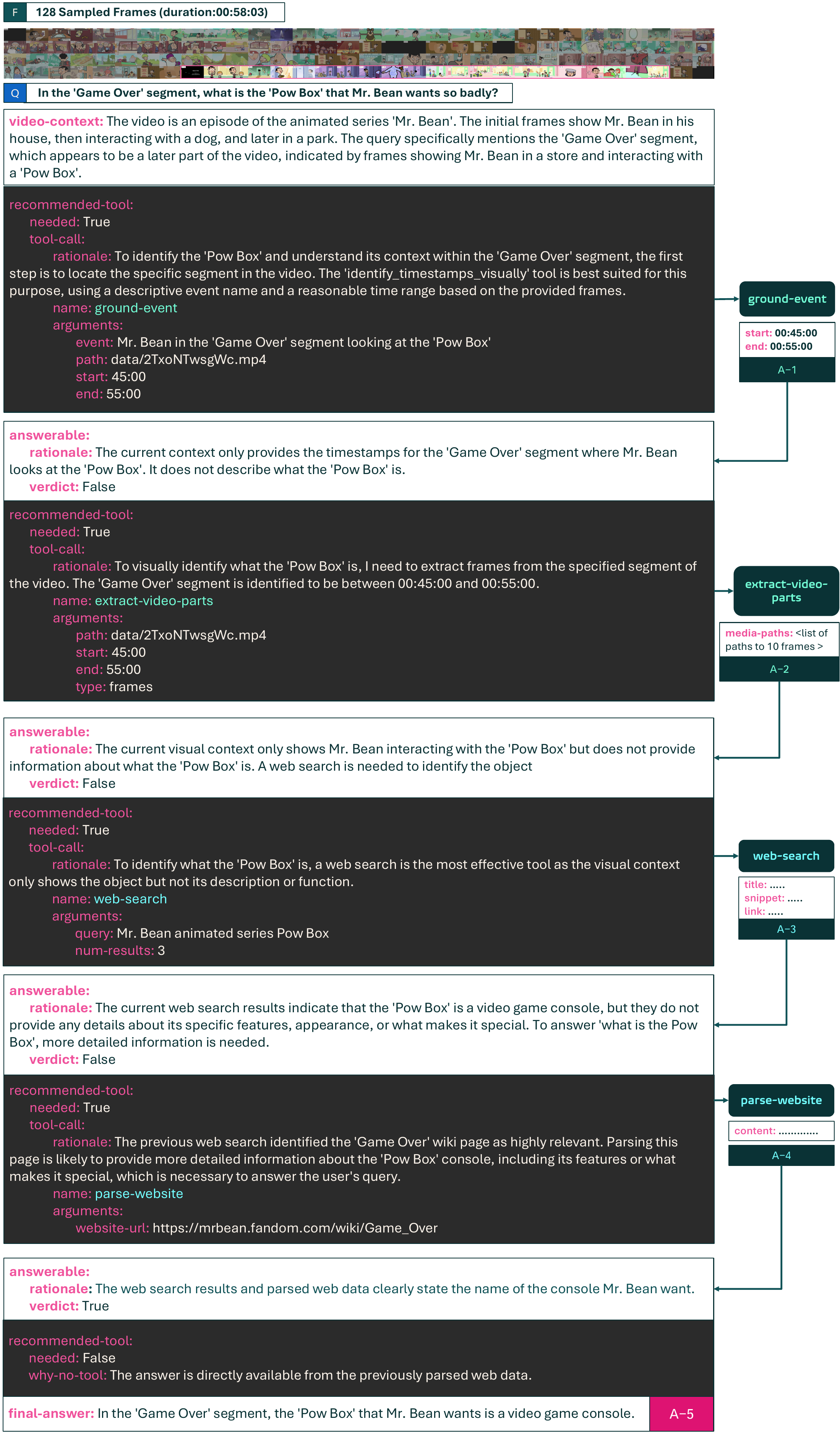} \\
\vspace{-0.1cm}
\caption{\textbf{\system Reasoning Trajectory Qualitative Example.} \system first coarsely locates the boundaries of the target segment given the sampled frames, followed by a web search to answer the question due to lack of information in the video. The {\direct} Qwen3-VL-8B-Instruct model (which has access to the full video transcript) predicts the following answer: \gr{\textit{The "Pow Box" is a fictional device in the "Game Over" segment, likely a humorous or exaggerated gadget that Mr. Bean desires, possibly for its comedic or absurd potential. The exact nature or function of the "Pow Box" isn't specified in the provided transcript, but it's implied to be something desirable or coveted by Mr. Bean, fitting his character's love for quirky, nonsensical objects.}}}
\label{fig:qual1}
\end{figure*}

\begin{figure*}[t!]
\centering
\includegraphics[width=0.8\linewidth]{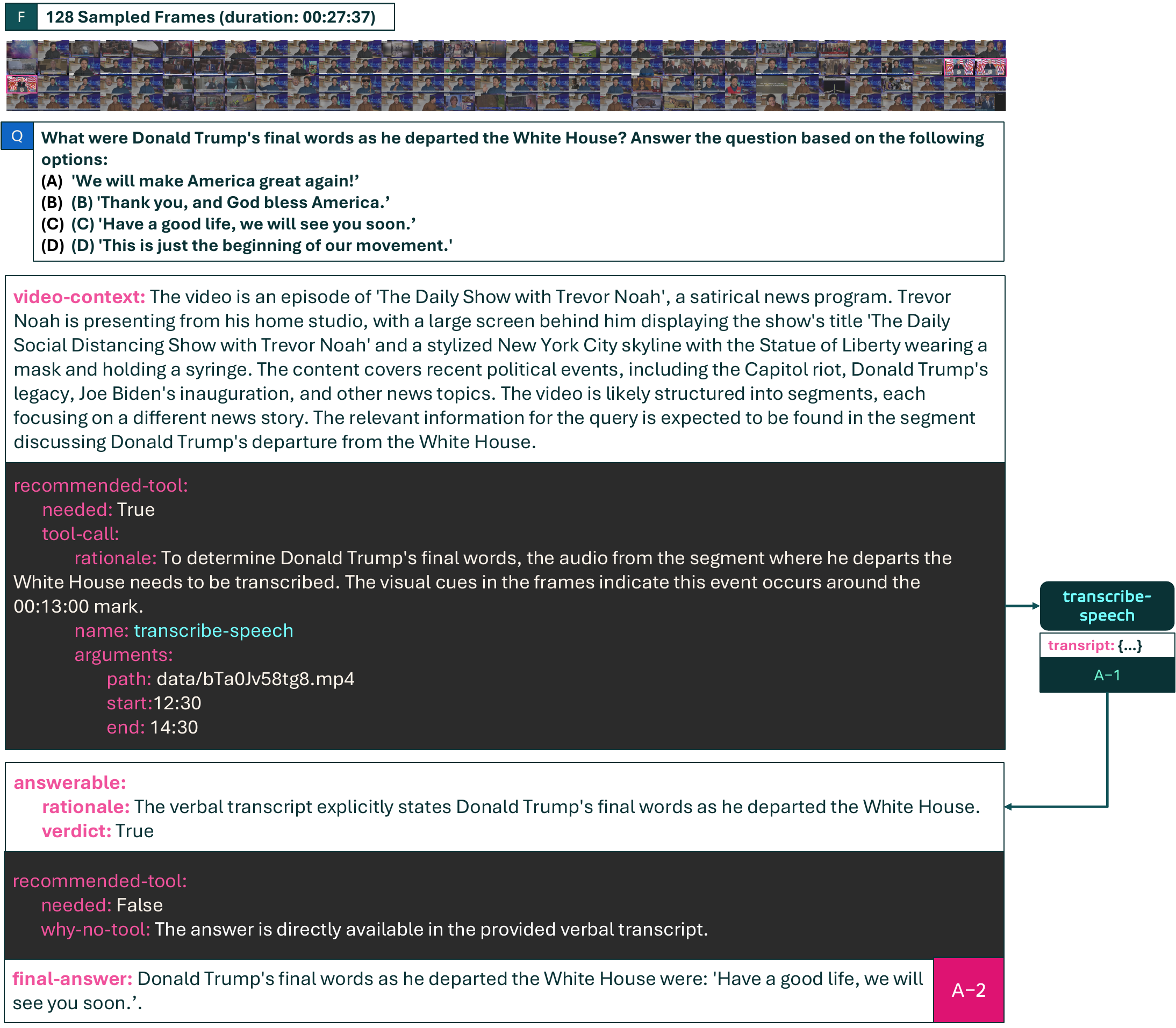} \\
\vspace{-0.1cm}
\caption{\textbf{\system Reasoning Trajectory Qualitative Example.} \system accurately transcribes only the target 2-minute segment to answer the user's question. The {\direct} Qwen3-VL-8B-Instruct model (which has access to the full video transcript) predicts the following answer: \gr{\textit{(B) 'Thank you, and God bless America.'.}}}
\label{fig:qual2}
\end{figure*}

\begin{figure*}[t!]
\centering
\includegraphics[width=0.75\linewidth]{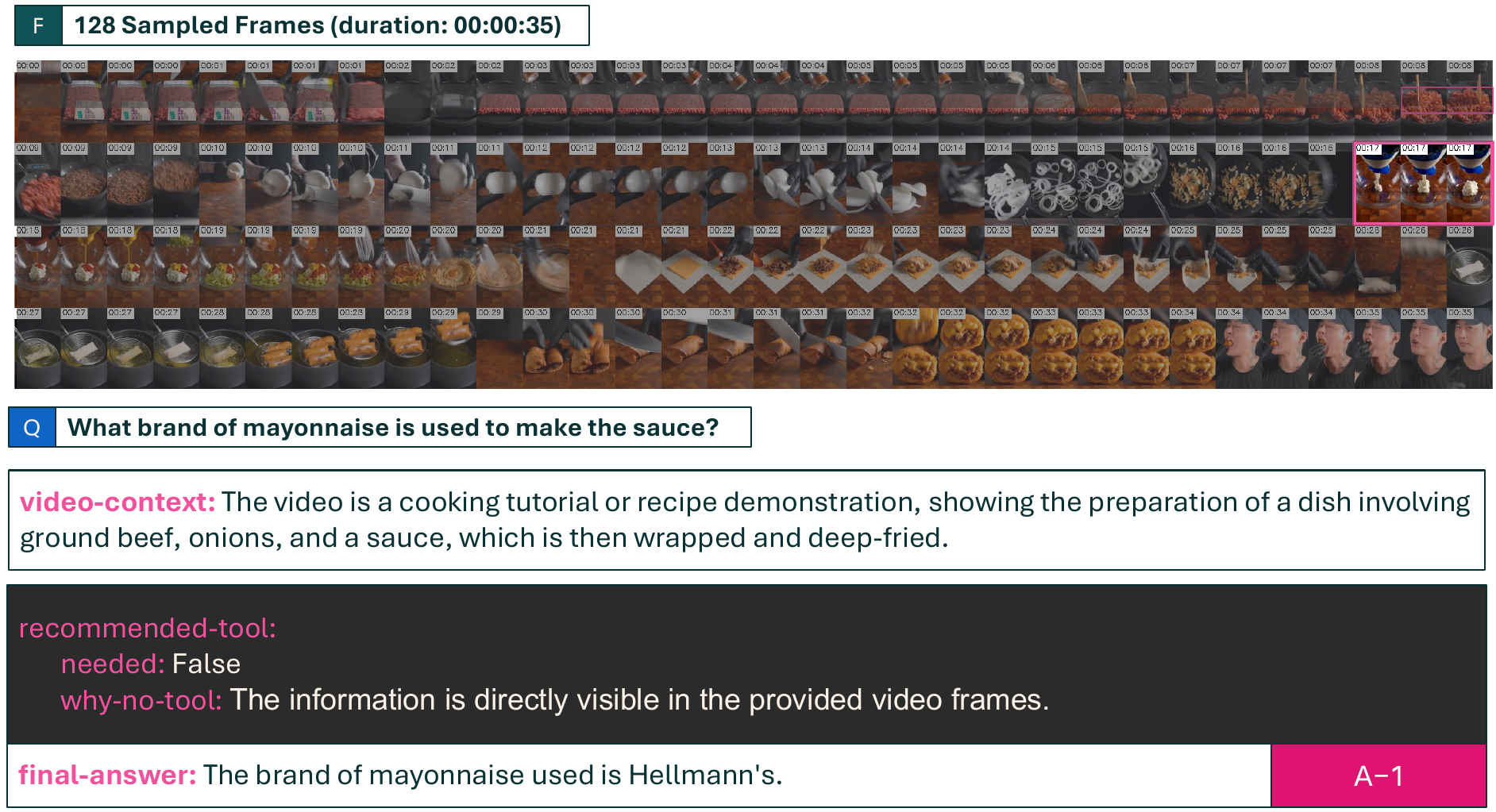} \\
\vspace{-0.1cm}
\caption{\textbf{\system Reasoning Trajectory Qualitative Example.} \system answers the question in a single turn for the short video. The {\direct} Qwen3-VL-8B-Instruct model (which has access to the full video transcript) predicts the following answer: \gr{\textit{Kraft}.}}
\label{fig:qual3}
\end{figure*}

\vspace{-0.4cm}
\paragraph{\system Any-Horizon Reasoning Trajectories.} We display qualitative examples to demonstrate the any-horizon reasoning abilities of \system in \cref{fig:qual1} (5 turns), \cref{fig:qual2} (2 turns), and \cref{fig:qual3} (single-turn).

{\thispagestyle{empty}}

\end{document}